\documentclass[11pt, a4paper,  copyright, gdm]{google}

\usepackage[authoryear, sort&compress, round]{natbib}
\uselogo{} 

\title{\thepa{} Paper Template}

\correspondingauthor{bohnetbd@google.com}

\renewcommand{\today}{2025-12-28}

\usepackage{amsmath,amsfonts,bm}

\def\eqref#1{equation~\ref{#1}}
\def\1{\bm{1}}

\DeclareMathAlphabet{\mathsfit}{\encodingdefault}{\sfdefault}{m}{sl}
\SetMathAlphabet{\mathsfit}{bold}{\encodingdefault}{\sfdefault}{bx}{n}

\newcommand{\fig}[1]{Figure \ref{#1}}
\usepackage[normalem]{ulem}

\usepackage{hyperref}
\usepackage{url}
\usepackage[utf8]{inputenc} % allow utf-8 input
\usepackage[T1]{fontenc}    % use 8-bit T1 fonts
\usepackage{hyperref}       % hyperlinks
\usepackage{url}            % simple URL typesetting
\usepackage{booktabs}       % professional-quality tables
\usepackage{amsfonts}       % blackboard math symbols
\usepackage{nicefrac}       % compact symbols for 1/2, etc.
\usepackage{microtype}      % microtypography
\usepackage{graphicx,color}
\usepackage{xspace}
\usepackage{subcaption}
\usepackage[ruled,vlined,linesnumbered]{algorithm2e}
\usepackage{float}

\usepackage{overpic}
\usepackage{wrapfig}
\usepackage{enumitem}
\usepackage{listings}
\usepackage{xcolor}
\usepackage{multirow}
\usepackage{algpseudocode}
\usepackage[flushleft]{threeparttable}
\usepackage{multicol}
\usepackage{fancyvrb}

\usepackage{booktabs}

\usepackage[most]{tcolorbox}

\newtcolorbox[auto counter, number within=section]{examplebox}[2][]{%
    enhanced,
    breakable,
    fonttitle=\bfseries,
    title=\thetcbcounter: #2,
    label={#1},
    colback=white,
    colframe=teal,
    colbacktitle=teal!85!black,
    coltitle=white
}

\usepackage{tikz}
\usepackage{placeins}

\title{Enhancing LLM Planning Capabilities through Intrinsic Self-Critique }

\usepackage{caption}
\author[1]{Bernd Bohnet}
\author[1]{Pierre-Alexandre Kamienny}
\author[1]{Hanie Sedghi}
\author[1]{Dilan Gorur}
\author[1]{Pranjal Awasthi}
\author[1]{Aaron Parisi}
\author[1]{Kevin Swersky}
\author[1]{Rosanne Liu}
\author[1]{Azade Nova}
\author[1]{Noah Fiedel}

\correspondingauthor{bohnetbd@google.com}

\affil[1]{\thepa{}{}}

\begin{abstract}

We demonstrate an approach for LLMs  to critique their \emph{own} answers with the goal of enhancing their performance that leads to significant improvements over established planning benchmarks. Despite the findings of earlier research that has cast doubt on the effectiveness of LLMs leveraging self critique methods, we show significant performance gains on planning 
datasets in the Blocksworld domain through intrinsic self-critique, without external source such as a verifier. We also demonstrate similar improvements on Logistics and Mini-grid datasets, exceeding strong baseline accuracies. 

We employ a few-shot learning technique and progressively extend it to a many-shot approach as our base method and demonstrate that it is possible to gain substantial improvement on top of this already competitive approach by employing an iterative process for correction and refinement. We illustrate how self-critique can significantly boost planning performance. 
Our empirical results present new state-of-the-art on the class of models considered, namely LLM model checkpoints from October 2024. 
Our primary focus lies on the method itself, demonstrating intrinsic self-improvement capabilities that are applicable regardless of the specific model version, and we believe that applying our method to more complex search techniques and more capable models will lead to even better performance.

\end{abstract}

\begin{document}

\maketitle

% \maketitle

\section{Introduction}

Recent advancements of Large Language Models (LLMs) have extended their applications to include planning, a domain traditionally dominated by algorithmic methods. LLM planning is important for a wide-range of tasks in which the task has implicit or explicit constraints and the LLM must generate a plan satisfying these constraints. Early investigations into LLM planning capabilities were unfavorable~\citep{valmeekam2023planning}, reinforcing the notion that "Language Models cannot plan". However, subsequent studies have introduced techniques, such as Many-Shot learning on planning tasks, to enhance these capabilities~\citep{agarwal2024manyshotincontextlearning, bohnet2024exploringbenchmarkingplanningcapabilities}.
Despite these enhancements, LLMs still lag behind classic planners that tackle algorithmically complex problems. Our approach introduces further improvements that are very promising.

For instance, where LLMs are typically tested on simplified  problems such as Blocksworld with 3-5~\citep{valmeekam2023planningabilitieslargelanguage} or 3-7 blocks~\citep{bohnet2024exploringbenchmarkingplanningcapabilities}, classic planners can be applied to considerably more complex problems. Nevertheless, there's a variety of natural-language tasks such as planning holiday trips or scheduling meetings \citep{geminiteam2024geminismall,hao2024largelanguagemodelsplan} which are  less computationally demanding, less classic planner friendly
and less structured (often being posed in natural language). These tasks are more difficult for classical planners compared to LLMs  underscoring the practical relevance of improving LLMs' planning capabilities. This range of tasks, from natural-language to classic planning tasks, highlights the importance of continuing to enhance LLMs' planning abilities, even though a gap remains with classic planners for more specialized, complex planning tasks. 

\begin{figure}[t]
    \centering
    \includegraphics[width=0.85\textwidth]{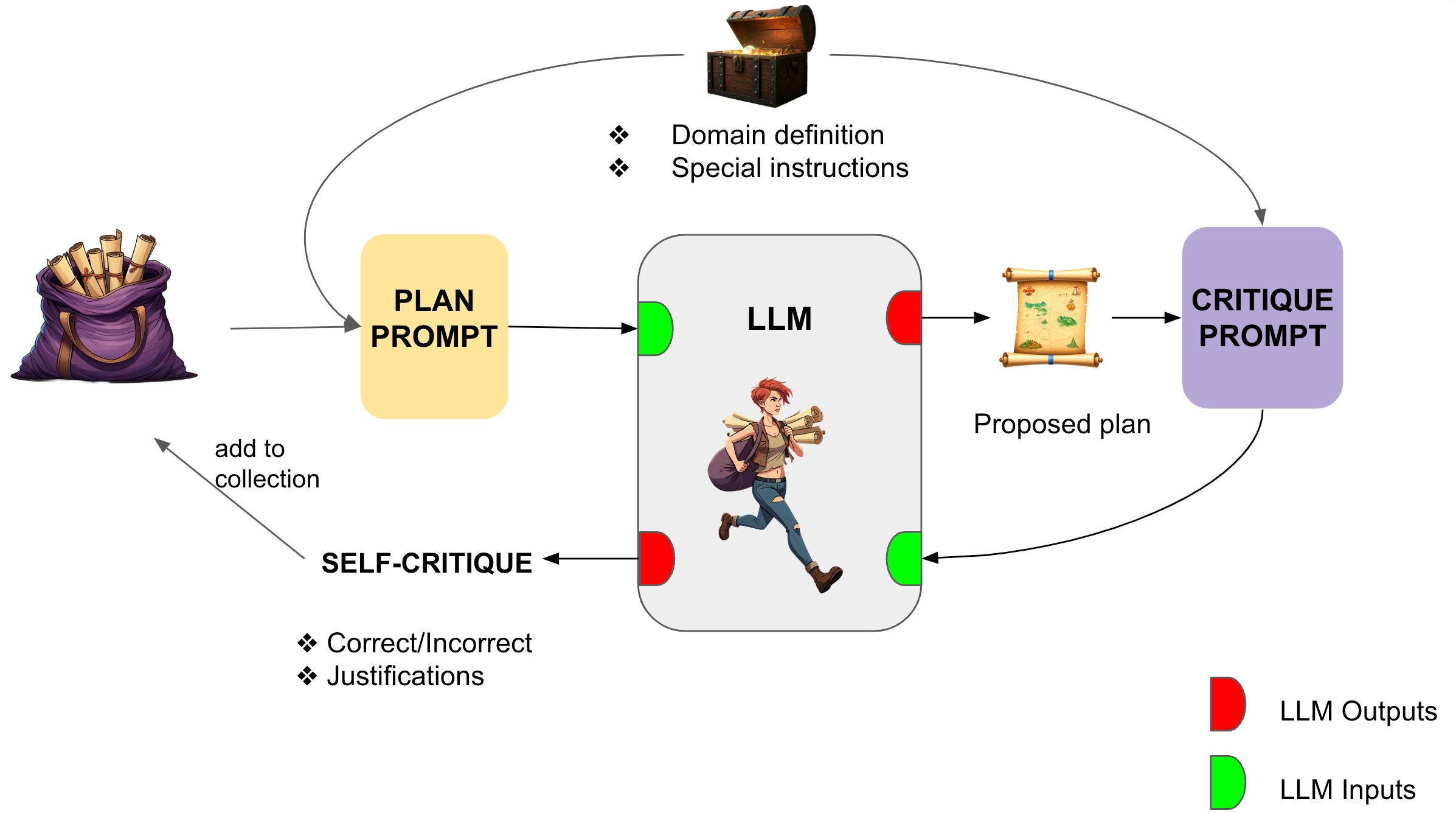}
    \caption{
    Illustration of the iterative self-improvement process for Large Language Models (LLMs) using in-context learning to incorporate self-critique feedback.
    The LLM, also represented by the explorer character, functions as the agent's brain, accepting prompts as inputs and generating outputs, represented by green and red semi-circles, respectively. Each iteration of the self-improvement mechanism comprises two key steps: i) plan generation and ii) self-critiquing, aimed at iteratively refining LLM outputs.
    In step i), the LLM generates a plan (symbolized by a map) based on a prompt incorporating domain-specific knowledge and instructions (symbolized by the treasure chest). Step ii) involves a self-critique mechanism where the LLM evaluates its own performance, providing correctness assessments and justifications, again leveraging domain knowledge.
    The process continues until a plan deemed correct is identified. Previous plans and their associated self-critique feedback are aggregated into a collection (symbolized by a bag), serving as contextual material for subsequent plan generation cycles.
    \protect\footnotemark}

    \label{fig:example}
    \label{fig:intro-figure}
\end{figure}
\footnotetext{Imagen 3  generated all pictures from Fig. \ref{fig:intro-figure}. 
}

The concept of self-improvement in LLMs (refining their responses based on self-generated feedback, often referred to as self-criticism), has gained significant attention often relying on an oracle for feedback \citep{yao2023tree,yao2023react,shinn2023reflexion}. 
This technique is particularly appealing in the context of planning tasks, where the ability to identify and rectify errors to improve the plans could lead to substantial performance gains. 

Earlier attempts to enhance LLM planning through self-critique yielded discouraging results, principally due to the lack of sufficient self-evaluation capabilities \citep{valmeekam2023largelanguagemodelsreally,huang2024largelanguagemodelsselfcorrect}.
The application of self-critique in prior work resulted in models that had high false-positive (FP) rates and that missed nearly all of the true negatives (TN), suggesting that current LLMs cannot effectively critique themselves \citep{valmeekam2023largelanguagemodelsreally}. 

Thus, earlier results on iterative plan improvements motivated the use of external feedback such as verification with an oracle (that has access to correct plan) or relying on human input to guide the process. 
%\aaron{Citation? Reference to discussion} This is also valmeekam but cited many times.
While these results give an indication of how much headroom there is for improving the plans in the presence of a perfect evaluator, the assumption that oracles are available at test time is unrealistic for most planning tasks of interest.

Exploring the limits of intrinsic LLM self-improvement - improving generations without the aid of external signals or further training - remains an active area of research \citep{huang2024largelanguagemodelsselfcorrect,singh2024scaling}. 
In this work, we propose an effective self-improvement method using zero-shot or few-shot prompts without the need for an external verifier. \fig{fig:intro-figure} illustrates the introduced self-improvement method via intrinsic self-critique in this paper, utilizing the LLM as the sole source of critique.
The Figure shows the iterative process of a plan generation followed by a self-critique step while adding previous failures as context (see Section \ref{sec:method}). The figure also highlights the crucial components: a clear definition of the domain and instructions (e.g. pre-conditions), proper prompt design, and the use of previous failures and critiques in subsequent plan refinement steps.

We investigate variations of our proposed method via ablation studies. We perform ablation study on different elements of our method.
Moreover, we explore how the method scales when across zero-shot and few-shot prompting and assess the impact of varying the number of self-critique iterations. We describe a novel approach to self-critique, where the LLM is prompted to evaluate the preconditions of each action and provided the state in a plan. Lastly, we examine the limitations of our method and evaluate the quality of self-criticisms aiming to provide a comprehensive understanding of the scalability and effectiveness of the proposed methods. 

Our focus lies on demonstrating intrinsic self-improvement capabilities of LLMs, independent of the specific model version. To this end, we use model checkpoints from October 2024 in our empirical studies. We conduct the exploratory experiments with Gemini 1.5 Pro \citep{gemini1-5} and confirm our results with other foundation models. Our work aims to demonstrate the applicability of our method to multiple foundational models, rather than comparing them to each other. 

In the Blocksworld domain, Gemini 1.5 Pro achieves significant performance gains on planning datasets. With the dataset from \cite{valmeekam2023planningabilitieslargelanguage} involving 3-5 blocks, we enhance performance from 49.8\% to 89.3\% through intrinsic self-critique, without external source such as a verifier. Similarly, for the dataset by \cite{bohnet2024exploringbenchmarkingplanningcapabilities} with 3-7 blocks, we boost Gemini 1.5 Pro accuracy from 57.2\% to 79.5\%. We also demonstrate similar improvements on Logistics and Mini-grid datasets, exceeding strong baseline accuracies.
In addition to results on Gemini 1.5 Pro, we have positive results with Claude 3.5 Sonnet \citep{claude} where accuracy improves from 68\% to 89.5\% and GPT-4o \citep{openai2024gpt4} for Blocksworld 3-5. 
With Gemma-2 27B \citep{gemmateam2024gemma2improvingopen} we only see a modest improvement for Logistics but not for Blocksworld, suggesting larger and more capable models are better able to self-improve.\footnote{For other models, see Table \ref{tab:llm-comparison}.}

The paper is organized as follows: Section \ref{sec:related-workd} reviews related work. Section \ref{sec:method} details the methods employed and Section \ref{sec:datasets} provides an overview on the datasets we use in our experiments. Section \ref{sec:experiments} presents the main findings across several datasets from related work, exploratory experiments on substantial number of validation examples to select hyper-parameters, and an in-depth investigation of what contributions of the self-critique method where important to improve the state-of-the-art. Finally, Section \ref{sec:conclusion} concludes the paper.

\section{Related Work}
\label{sec:related-workd}

Our study builds upon the findings of \citet{valmeekam2023largelanguagemodelsreally}, who offered a critical assessment of the self-critiquing capabilities LLMs for planning. Contrary to more optimistic views on the self-correct, self-critique, reflexion, or iterative self-refinement abilities of LLMs \citep{madaan2023selfrefine,welleck2023generating,shinn2023reflexion},
for the Blocksworld eval, \citet{valmeekam2023largelanguagemodelsreally} observed a modest improvement when relying exclusively on LLMs for both generation and verification. However, they highlight a critical issue: the use of LLMs for verification leads to a significant number of false positives.

\citet{huang2024largelanguagemodelsselfcorrect} explore whether modern LLMs can self-correct their reasoning, concluding they cannot reliably self-correct. They introduce the concept of {\em intrinsic self-correction} and find that despite prevailing optimism, LLMs struggle with this ability, as shown through experiments involving multiple prediction and self-correction steps that increase latency due to dual calls to the LLM. 

Their analysis shows that performance varies across models and datasets; for instance, GPT-4 and Llama-2 both experience performance declines on the GSM8K and CommonSenseQA datasets, with the extent of decline linked to the model's intrinsic reasoning capabilities as well as prompt design. They also observe that attempts at self-correction can lead to correct outcomes being made wrong or inaccuracies being maintained through the reasoning chain. Stronger models, however, tend to show less performance drop.

%\citet{huang2024largelanguagemodelsselfcorrect} emphasizes further the importance of careful experimental design in future self-correction research, including fair comparisons and proper prompt design. The focus of \cite{huang2024largelanguagemodelsselfcorrect} is on the challenges in self-improvement, though they acknowledge that self-correction might be successful in other domains or future models.

The focus of our work is on self-correction for planning, whereas \citet{huang2024largelanguagemodelsselfcorrect} concentrate on the challenges of self-correction in reasoning.
Planning tasks differ from reasoning, as they have a set of constraints and "valid states" provided explicitly or implicitly by the problem definition. For classical reasoning tasks, the space of states, actions, and preconditions tend to be explicitly provided in the domain definition. Hence the verification of a LLM output can follow defined rules by checking the preconditions of each action and providing a follow up state. 

\section{Benchmarks}
\label{sec:datasets}

In this work, we focus on planning tasks utilizing a formal language based on Planning Domain Definition Language (PDDL)~\citep{McDermott1998PDDLthePD} and keep the symbolic representation of the problem / action / state definition specified by PDDL. Previous research has demonstrated that the planning performance of LLMs on Natural Language and PDDL is comparable using few-shots~\citep{bohnet2024exploringbenchmarkingplanningcapabilities}. 

For each instance in these datasets, a problem is given and a solution plan is requested. A response/plan is considered \emph{correct} when the goal is reached; otherwise, it is counted as \emph{incorrect}. 

For our experiments, we selected datasets that have been widely utilized in prior works~\citep{valmeekam2023planning, bohnet2024exploringbenchmarkingplanningcapabilities, stein2024autoplanbench} to ensure meaningful comparisons between our method and existing results.

%Incorrect plans corresponds to two cases: i) invalid plans, i.e. the plan is not valid or the goal is not reached, ii) successful plans, i.e. the goal is reached.

\paragraph{Classical Planning Benchmarks} These benchmarks are all from AI planning problems one can generate as hard or easy and as many instances given the sample generation code.
\begin{enumerate}
    \item \textbf{Blocksworld} consists a number of blocks on a table and a hand to move them from an initial configuration to another.
    \item \textbf{Logistics} includes planning deliveries of packets between cities using trucks and planes.
    \item \textbf{Minigrid} includes various rooms with random configurations and keys where a robot needs to navigate from an initial position to a final position.
\end{enumerate}

\paragraph{Mystery Blocksworld} This dataset is inspired by a mystery dataset in planning domain and is similar to the one in~\citep{valmeekam2023planningabilitieslargelanguage} in nature. This dataset is notable for its obfuscation of actions and attributes into non-specific or deceptive predicates. We randomly choose some samples and replace some words of description, action, predicates or object names either with non-specific or deceptive which remain consistent throughout the dataset. The goal of such design is to see if models are using some common-sense knowledge in coming up with plans and how much obfuscation of this information hurts the model prediction. 
In the results we report the experiments section here, we use deceptive version. We also ran experiments with the non-specific counterparts and gained impressive results. 

%\paragraph{AutoPlanBench} %\hanie{pointer for pranjal}

\paragraph{AutoPlanBench}

We also compare our propose method on the planning problem domains from the AutoPlanBench \citep{stein2024autoplanbench} repository. AutoPlanBench contains 10 new domains in addition to the ones in \citep{valmeekam2023planning} with 21 problems per domain. We ignore the {\em floortile} domain as no existing methods (including ours) achieve any non-trivial accuracy for this particular domain. For a given domain, each problem is represented in both PDDL and natural language and also comes optimal golden plans. For each planning domain, the optimal plan length varies between 3 and 20 steps. As mentioned before, in our experiments we make use of the PDDL representations of the problems. 

{\bf Datasets.} Further, we drew upon the Blocksworld dataset described in \cite{bohnet2024exploringbenchmarkingplanningcapabilities}, which features problems with 3-7 blocks, consisting of a training set, validation and test sets. The validation and test set each contain 1000 problems. In addition to these, we employed the Logistics and Mini-Grid datasets, each with its own test set of 600 problems \citep{bohnet2024exploringbenchmarkingplanningcapabilities}.
% Further, we use the planning datasets from \citep{stein2024autoplanbench}. The authors release 9 new domains in addition to the ones in \citep{valmeekam2023planning} with 21 problems per domain, and we apply our method to the provided data in the provided PDDL format. 

Specifically, we incorporated datasets from \citep{valmeekam2023planning}, which consist of problems from Blocksworld dataset  involving 3-5 blocks across a set of 600 problems, in addition to the mystery Blocksworld dataset. The latter dataset is notable for its obfuscation of actions and attributes into non-specific predicates, also encompassing 600 problems (see Appendix \ref{app:sample-trace-my} for domain definition and example problem).

\section{Method}
\label{sec:method}

The self-critique methodology studied in this work involves an iterative refinement of responses from the LLM until a correct solution is found, as determined by the LLM's own assessment. The iterative procedure is described as in Algorithm \ref{alg:self-critique} and is explained below. 

\begin{algorithm}[th]
\caption{Self-Critique Algorithm}
\label{alg:self-critique}
\small
\SetAlgoLined
\SetAlgoNoEnd
\begin{multicols}{2}
\SetKwInput{KwMain}{Main Algorithm}
\KwMain{}
\KwIn{Problem definition $\mathcal{D}$, maximum iterations $k$, self-consistency samples $N$, a LLM sampling distribution $P$ is parametrized by $P(p, n)$ where $p$ is the context and $n$ number of samples}
\KwOut{Final plan}
\setlength{\columnseprule}{0.4pt}
\SetKwFunction{PlanGeneration}{PlanGeneration}
\SetKwFunction{SelfCritique}{SelfCritique}
\SetKwFunction{Revise}{Revise-Prompt}
$\tau \gets \emptyset$\;
\For{$step = 0$ \textbf{to} $k$}{
    $plan \gets$ \PlanGeneration{$\mathcal{D}, \tau$}\;
    $critique \gets$ \SelfCritique{$\mathcal{D}, plan, N$}\;
    \If{$critique$ deems $plan$ correct}{
        \textbf{break}\;
    }
    $\tau \gets$ \Revise{$\tau, plan, critique$}\;
}
\KwRet $plan$

\columnbreak
\SetKwInput{KwHelp}{Helper Functions}
\KwHelp{}
\SetKwProg{Fn}{Function}{:}{}
\Fn{\PlanGeneration{$\mathcal{D}, \tau$}}{
    $prompt \gets$ PlanPrompt($\mathcal{D}, \tau$)\;
    \KwRet $P$($prompt, 1$)\;
}
\BlankLine
\Fn{\SelfCritique{$\mathcal{D}, plan, N$}}{
    $prompt \gets$ CritiquePrompt($\mathcal{D}, plan$)\;
    \KwRet Self-Consistency($P$($prompt, N$))\;
}
\BlankLine
\Fn{\Revise{$\tau, plan, critique$}}{
    \KwRet $\tau \oplus plan \oplus critique$ \;
}
\end{multicols}
\end{algorithm}

For $step$ from $0$ to $k$, repeat:
\begin{enumerate}
    
    \item {\bf Plan generation (lines 9-11):} A LLM, also represented by the explorer character in Fig.~\ref{fig:intro-figure},  receives a prompt to solve a given planning problem and generates a $plan$ from some distribution $P$. The prompt includes the previous predicted solutions and self-critique outputs from previous iterations (for $k>1$). If the prompt exceeds a predefined length limit, the loop is exited by returning the plan from the previous step.
    The plan generated by the explorer is represented as a map in  Fig.~\ref{fig:intro-figure}.

    \item {\bf Self-Critique (lines 12-14): } The same LLM is prompted to generate a critique of its proposed plan, with a prompt that includes both the problem definition $\mathcal{D}$ and its predicted $plan$. We optionally apply Self-Consistency~\citep{wang2023selfconsistency_cot} which uses majority voting. 
    The explorer is capable of reflecting on its own predictions! 
    \item {\bf Revise (lines 15-16): } If the LLM deems the response from the self-critique as correct (line 5), the process concludes (line 6, equivalently to early stopping), and the final plan is returned, thereby exiting the loop. 
    Otherwise, add the plan to the prompt (line 16) and go back to planning step (line 3). The LLM adds failures to its failure collection, represented by the bag in Fig.~\ref{fig:intro-figure}.
%B: done    \kevin{Reminder to change this} 
\end{enumerate}

% Algorithm \ref{alg:self-critique} formally presents the method outlined above. The enumerated steps are (1) Plan, (2) Critique, and (3) Revise executed in a loop (line 2-7), leveraging the LLM's capabilities to iteratively refine the  solution until the critic deems the plan correct (line 5).

% In this work, we focus on planning tasks utilizing a formal language based on Planning Domain Definition Language (PDDL)~\citep{McDermott1998PDDLthePD} and keep the symbolic representation of the problem / action / state definition specified by PDDL. Previous research has demonstrated that the planning performance of LLMs on Natural Language and PDDL is comparable using few-shots~\citep{bohnet2024exploringbenchmarkingplanningcapabilities}. 
%Note that we include ablations on the impact of PDDL on our self-critique method in Section \ref{sec:discussion}.
%,allowing us to bypass the complexities associated with converting from formal definitions to Natural Language and back. 

%for the prompt example and related work for the many-shot technique \citep{agarwal2024many,bohnet2024exploringbenchmarkingplanningcapabilities}.
This allows us to investigate the behavior of the self-critique method, on top of a planning process which scales favorably with the number of few-shot examples. The prompts are randomly sampled from the respective training set in a deterministic manner, ensuring that the few-shot exemplars for the same problem remain consistent across ablations. This approach facilitates more reliable comparisons between experiments.

{\bf Self-Critique prompt.} The design of the self-critique prompt is crucial for accuracy. %, as discussed in Section \ref{sec:discussion}. 
It includes: (1) the domain definition $\mathcal{D}$ with actions, their preconditions, and effects. Notably, including the preconditions and a description of how the succeeding state is created are essential for a consistent verification process (2) Optionally, we provide few-shot exemplars for verification. (3) Instructions on how to verify each action in-sequence are also included. 
Section \ref{sec:discussion} demonstrates how these elements collectively improve our method's performance. In most experiments, we use zero-shot self-critique prompt templates (Appendix \ref{app:self-eval-prompt-no-few-shot}) and for Blocksworld, an 8-shot self-critique prompt template is also employed (Appendix \ref{app:self-eval-prompt-few-shot}).
We bootstrap the few-shot self-critique prompts from a single handcrafted example and obtain outputs by running the self-critique method on the training set. We ensure that the selected prompt ratings (correct/wrong) are consistent with a PDDL validator.

\section{Experiments}
\label{sec:experiments}

Our experiments assess the performance of Self-Critique methods on planning benchmarks, specifically on PlanBench \citep{valmeekam2023largelanguagemodelsreally}, AutoPlanBench \citep{stein2024autoplanbench} and the Planning-Capabilities  \citep{bohnet2024exploringbenchmarkingplanningcapabilities}. 
We apply our method described in Section~\ref{sec:method}; iteratively prompting the model and having it critique its own predictions. This process continues until either the LLM deems the solution correct or it exceeds some predefined number of iterations. 
In all exploratory experiments, we use Gemini 1.5 Pro \citep{gemini1-5}, GPT-4o \citep{openai2024gpt4}, Claude 3.5 Sonnet \citep{claude}, as well as Gemma-27b \citep{gemmateam2024gemma2improvingopen} as an open source alternative (Appendix \ref{ap:gemma_results}).
The last proposed plan in the iterative process is used for computing the accuracy and it is validated by using a PDDL-validator VAL\footnote{We use the VAL-PDDL verifier available from \url{https://github.com/KCL-Planning/VAL}}. 

In Subsection \ref{sec:main-results}, we present the main results of this approach and compare them with state-of-the-art results from the above planning benchmarks. In Subsection \ref{sec:setup-and-exploration}, we describe exploratory experiments on the validation sets. Finally in Section \ref{sec:discussion}, we shed more light on the main findings and provide further analysis and an ablation study.  

\subsection{Main results}
\label{sec:main-results}

\begin{figure}
    \centering
        \includegraphics[width=\textwidth]{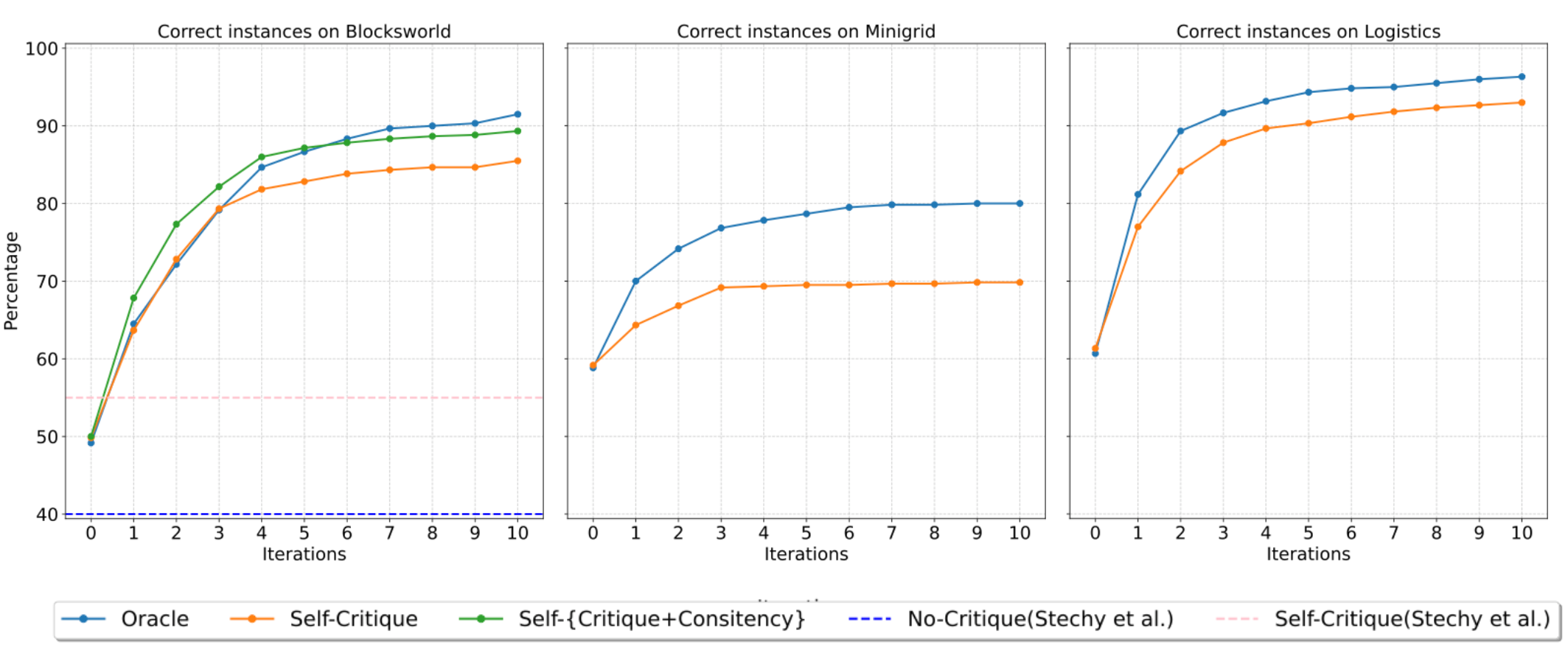}
    
    \caption{
The figure illustrates how number of correct instances increases as we increase the number of Self-Critique iterations on Blocksworld (left), Mini-Grid (center) and Logistics (right). The initial step (step $0$), depicted in the figure, represents the baseline  without Self-Critique. Subsequent iterations, from 1 to 10, each involve a Self-Critique and a refinement request to an LLM, demonstrating progressive improvements of correct instances. We use a $16$-shot planning prompt and $0$-shots for the critique prompt on all benchmarks. Model: Gemini 1.5 Pro
%\pierre{@bernd need precision here: 1) temp with self-consistency does not make sense, 2) why is the file path for minigrid and blocksworld se16e8 ? does that suggests 8 critic shots?}
    }
    \label{fig:stepwise-accuracy}
\end{figure}

Figure~\ref{fig:stepwise-accuracy} depicts the impact of our Self-Critique method on accuracy  as the number of steps increases. The model is Gemini 1.5 Pro. We observe that across all three benchmarks our Self-Critique method dramatically increases the accuracy across multiple steps.
A large portion of the improvements on Blocksworld, Minigrid and Logistics happen during the first iteration of self-improvement, but we note that the accuracy keeps improving with the number of iterations. 
We compare Self-Critique methods against the accuracy of the Oracle, which uses the validator for feedback in the self-improvement process, thus eliminating the error due to inaccuracy of Self-Critique, testing the iterative planning capability.
Lastly, we  plot Self-\{Critique+Consistency\}, i.e. Self-Critique with Self-Consistency, where the final outcome is determined by aggregating the results of 5 independent Self-Critique calls and selecting the outcome with the most votes. This is the green line in the Blocksworld subfigure.
We only applied it to Blocksworld for compute budget reasons and observed that the gap between Self-\{Critique+Consistency\} and the Oracle is really small, showing that the evaluation capabilities are substantially improved by using Self-Consistency.

% We see a big improvement in the first few steps, and although the gains get smaller, we inch towards perfect accuracy.
% Bernd: allowing the figure to float and removed [H]
% as it causes huge gaps of empty space 

\begin{figure}
    \centering
    \begin{subfigure}[b]{0.33\textwidth}
        \centering
        \includegraphics[width=\textwidth]{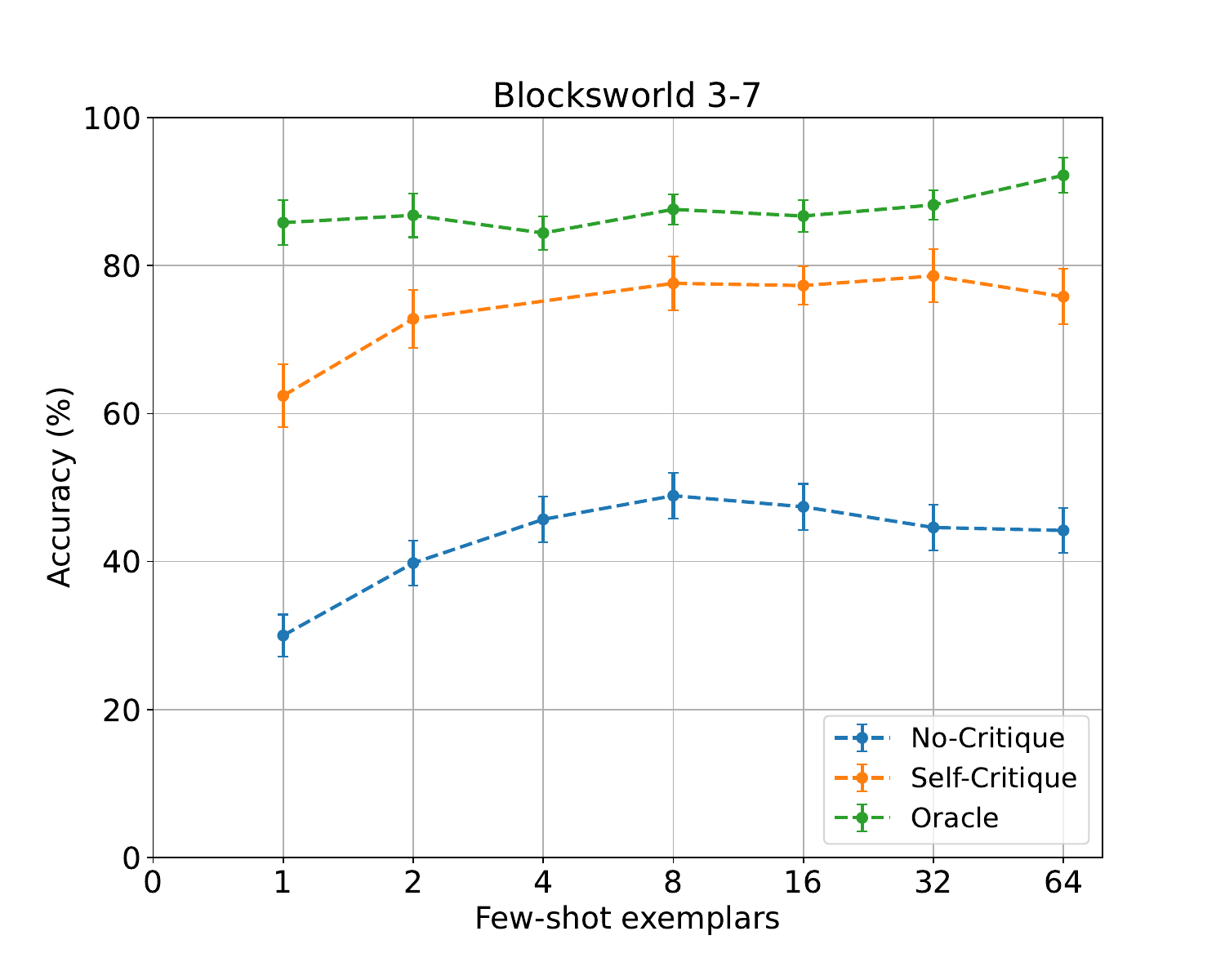}
        \label{fig:bw}  
    \end{subfigure}%
    \hspace{-0.5em}%
    \begin{subfigure}[b]{0.33\textwidth}
        \centering
        \includegraphics[width=\textwidth]{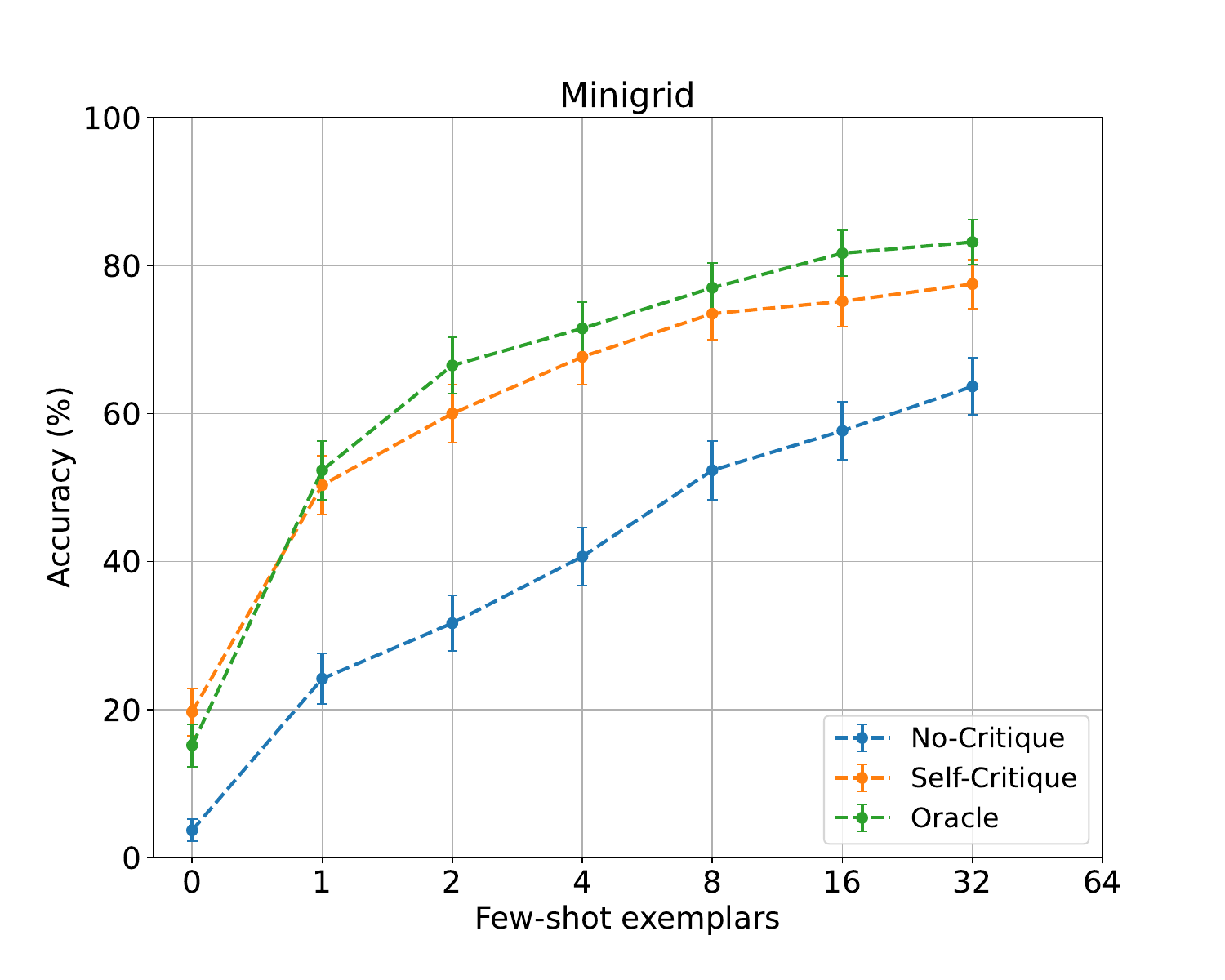}
        \label{fig:mg}  
    \end{subfigure}%
    \hspace{-0.5em}%
    \begin{subfigure}[b]{0.33\textwidth}
        \centering
        \includegraphics[width=\textwidth]{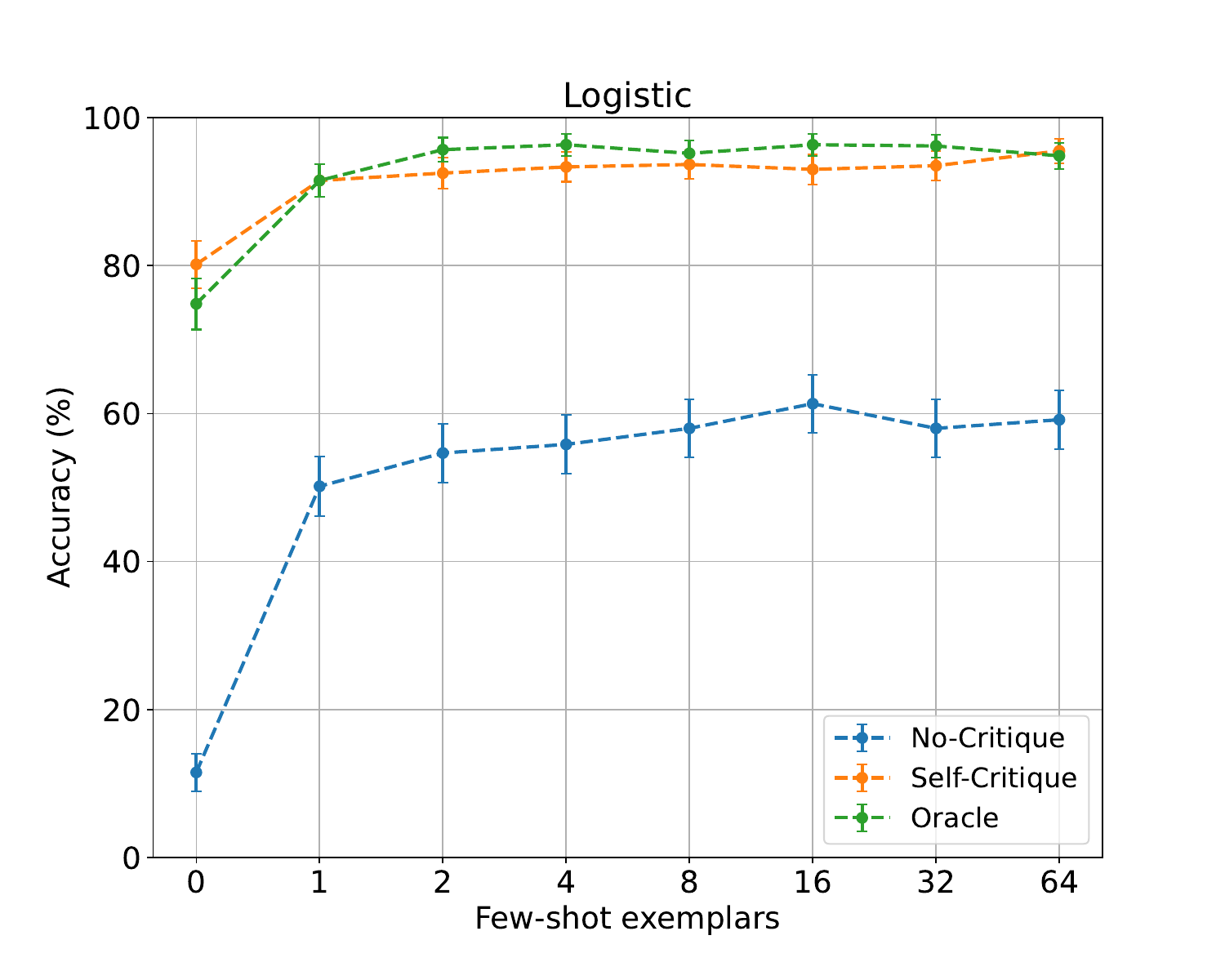}
        \label{fig:lg}  
    \end{subfigure}
    \caption{Performance on Blocksworld (left), Mini-Grid (center) and Logistics (right) with increasing number of shots in the planning prompt. While performance on Logistic stagnates using more than 2 shots, the more exemplars in Blocksworld and Minigrid the better.
    To strike a good balance between LLM context length and accuracy over all benchmarks, we used this experiment to select the number of planning shots, and chose it to be 16 exemplars. Model: Gemini 1.5 Pro} 
    \label{fig:num-shots-examples} 
\end{figure}

Figure~\ref{fig:num-shots-examples} illustrates the improvements achieved by employing an increasing number of few-shots exemplars as the baseline method, enhanced by the addition of the Self-Critique method. 
The graphs depict curves with an increasing number of shots up to 64, with curves representing No-Critique (orange), Self-Critique (pink), and Oracle (blue). The 0-shot setting contains only the domain definition without any additional exemplars. The model used is Gemini 1.5 Pro.

% presents the results using datasets from \cite{bohnet2024exploringbenchmarkingplanningcapabilities}.
% Table \ref{tab:logistics-minigrid-bw-datasets} shows results on five additional datasets from \citet{bohnet2024exploringbenchmarkingplanningcapabilities} and demonstrates the effectiveness of the Self-Critique method on other datasets. The first logistics dataset, with up to four places and two packages, can be nearly solved using Self-Critique techniques, reaching 93.2\% accuracy. Meanwhile, a more challenging logistics dataset with up to 4 cities, 2 places per city and 8 packages, remains overall difficult, but Self-Critique improves the results substantially, from 18.9\% to 32.8\%. For Minigrid, we demonstrate significant gains.
% For Blocksworld 3-7, the gains are similar to those in Blocksworld 3-5, but all results are at a slightly lower level, probably due to the increased complexity of the solutions as we increase the number of blocks.

Table \ref{tab:sota-comparison} shows in addition the final results along with results reported from \cite{stechly2024chain} for Blocksworld 3-5. As \cite{stechly2024chain} use GPT-4 and Natural Language for planning, we investigate the influence of these settings in Subsection \ref{sec:discussion} (see Table \ref{tab:natural_language_pddl}).
\begin{table}
\caption{Accuracy comparisons across various methods for the Blocksworld 3-5 dataset, with results taken from \cite{stechly2024selfverification} (left). All accuracy values are expressed in percentages (\%) and include the computation of 95\% Confidence Intervals. Our method achieves SoTA results by utilizing 8-shot self-critic prompts, as detailed in Appendix \ref{app:self-eval-prompt-few-shot} and self-consistency (5 votes). }

\label{tab:sota-comparison}
\begin{center}
\setlength{\tabcolsep}{3pt} % default is 6pt
\begin{tabular}{lrrrcccccc}
\multicolumn{1}{c}{} & \multicolumn{3}{c}{\bf \citep{stechly2024selfverification}} &  \multicolumn{4}{c}{\bf This work} \\
\cmidrule(lr){2-4} \cmidrule(lr){5-6}  \cmidrule(lr){5-8}

% \hline
     & {\bf No-Critique} & {\bf Critique} & \multicolumn{1}{c}{\bf Oracle}  & 
{\bf No-Critique} &  \multicolumn{1}{c}{\bf Critique } & \multicolumn{1}{c}{\bf Critique+SC} & \multicolumn{1}{c}{\bf Oracle}  \\

\hline 
Blocksw. 3-5 & 40 & 55 & 87  & 49.8$\pm$4.0 & 85.5$\pm$2.8 & 89.3$\pm$2.5   &91.5$\pm$2.3    \\
Mystery BW & 4 & 0 & 8  &22.3$\pm$3.3 & 35.2$\pm$3.8 & 37.8$\pm$3.9 &37.3$\pm$3.9\\ % 
\end{tabular}
\end{center}
\end{table}

Table \ref{tab:logistics-minigrid-bw-datasets} shows results on five additional datasets from \citet{bohnet2024exploringbenchmarkingplanningcapabilities} and demonstrates the effectiveness of the Self-Critique method on other datasets. The first logistics dataset, with up to four places and two packages, can be nearly solved using Self-Critique techniques, reaching 93.2\% accuracy. Meanwhile, a more challenging logistics dataset with up to 4 cities, 2 places per city and 8 packages, remains overall difficult, but Self-Critique improves the results substantially, from 18.9\% to 32.8\%. For Minigrid, we demonstrate significant gains.
For Blocksworld 3-7, the gains are similar to those in Blocksworld 3-5, but all results are at a slightly lower level, probably due to the increased complexity of the solutions as we increase the number of blocks.

\begin{table}[th!]
\caption{
Results on additional planning datasets demonstrate the effectiveness of the method across diverse problems. Model: Gemini 1.5 Pro. Datasets marked with $*$ are sourced from \cite{bohnet2024exploringbenchmarkingplanningcapabilities}. In these experiments, a zero-shot self-critic is employed, which incorporates the domain definition (see detailed prompt in the Appendix \ref{app:self-eval-prompt-no-few-shot}). The Logistic and Mini-Grid Hard datasets contain more complex tasks. }
\label{tab:logistics-minigrid-bw-datasets}
\begin{center}
\begin{tabular}{lrrrrrrrr}

% {\bf Method} & \multicolumn{1}{c}{\bf few-shot} & {\bf few-shot} & {\bf few-shot} &  \\

% \hline
{\bf Dataset} & {\bf No-Critique} & {\bf Self-Critique} & \multicolumn{1}{c}{\bf Oracle} &  \\
% \hline \\
\hline 
Logistics$*$   & 60.7$\pm$3.9 & 93.2$\pm$1.8  & 95.0$\pm$1.7 & \\
Logistics Hard & 18.9 $\pm$2.4 & 32.8$\pm$2.9 & 38.8$\pm$3.9 \\
% Mini-Grid$*$   &  59.2$\pm$3.9 & 69.8$\pm$3.6 & 79.8$\pm$3.2  &   \\ old with shorter context
Mini-Grid$*$   &  57.7$\pm$3.9 & 75.2$\pm$3.5 & 79.8$\pm$3.2  &   \\

Mini-Grid Hard &  39.7$\pm$4.0 & 43.5$\pm$4.0 & 52.3$\pm$4.0 &   \\
Blocksworld$*$  3-7 & 57.2$\pm$3.1  & 79.5$\pm$2.5 & 92.7$\pm$1.6    \\
\end{tabular}
\end{center}
\end{table}

Table \ref{tab:one-shot-auto-plan-bench} also presents results on  AutoPlanBench~\citep{stein2024autoplanbench}, which features a number of datasets across various problem types, each containing only 21 instances. In line with our other experiments, we employ a few-shot methodology, utilizing just a single shot and augmented by Self-Critique technique. We observe overall improvements across all datasets as a result of applying Self-Critique.
%Bernd this is not entirely correct as we did not evaluate the performance, I suggest to leave this as it is.
%We evaluate the performance of the Act agent, denoted as Oracle(Act) in the table,  corresponding to an Oracle feedback at every step, an even stronger assumption than Oracle feedback at the end of plan generation. 

\begin{table}
\centering
\caption{AutoPlanBench with 1-shot plan and zero-shot Self-critique results and  Act and CoT taken from \citep{stein2024autoplanbench}. These experiments demonstrate performance enhancements across additional datasets. However, with only 21 exemplars available per task, more robust few-shot approaches were not investigated. Note that the Act method uses (per step) golden feedback from the validator and hence is not directly comparable with the proposed Self-Critique method.
%\pierre{moved Act to last column has it is the same as Oracle but for every step. @Pranjal can you complete Oracle column?}
}

%\pierre{i am not sure about 1-shot here. what is the difference with BASE?}

\label{tab:one-shot-auto-plan-bench}
\begin{tabular}{lrrrr}
\textbf{Dataset} & \textbf{Act} & \textbf{CoT} & \textbf{1-shot} & \textbf{Self-Critique} \\
\hline
Goldminer & 30 & 20 & 2.8 $\pm2.3$& 32.3 $\pm$9.3\\
Rovers           & 50  & 10  & 0$\pm$0.0      & 7.6$\pm$6.4  \\
Grid             & 70  & 20  & 9.5$\pm$0.0    & 53.3$\pm$7.6 \\
Grippers         & 55  & 75  & 25.7$\pm$5.7   & 79.0$\pm$6.4 \\
Satellite        & 90  & 50  & 4.7$\pm$3.0    & 91.4$\pm$4.6 \\
Depot            & 20  & 15  & 0$\pm$0.0      & 20.9$\pm$3.8 \\
Movie            & 100 & 100 & 3.8$\pm$1.9    & 100$\pm$0.0  \\
Ferry            & 40  & 95  & 44.8$\pm$8.8   & 61.9$\pm$6.0 \\
Vistall          & 100 & 85  & 96.1$\pm$3.5   & 99.0$\pm$1.9 \\
\end{tabular}
% \begin{tabular}{lrrrrr}
% \textbf{Dataset} & \textbf{CoT} & \textbf{No-Critique} & \textbf{Self-Critique} & \textbf{Oracle} & \textbf{Oracle(Act)} \\
% \hline
% Rovers           & 10  & 0$\pm$0.0      & 7.6$\pm$6.4  && 50  \\
% Grid             & 20  & 9.5$\pm$0.0    & 53.3$\pm$7.6 && 70  \\
% Grippers         & 75  & 25.7$\pm$5.7   & 79.0$\pm$6.4 & &55  \\
% Satellite        & 50  & 4.7$\pm$3.0    & 91.4$\pm$4.6 && 90  \\
% Depot            & 15  & 0$\pm$0.0      & 20.9$\pm$3.8 && 20  \\
% Movie            & 100 & 3.8$\pm$1.9    & 100$\pm$0.0  && 100 \\
% Ferry            & 95  & 44.8$\pm$8.8   & 61.9$\pm$6.0 && 40  \\
% Vistall          & 85  & 96.1$\pm$3.5   & 99.0$\pm$1.9 && 100 \\
% \end{tabular}
\end{table}

% Additionally, we include detailed error analysis from the oracle, outlining changes to state and the reasons and actions at which the validation fails.
% \pierre{@bernd complete if we have some cool qualitative analysis on non physical plans}

\subsection{Setup Exploration}
\label{sec:setup-and-exploration}

%The method has a number of parameters and possible setups: for both the plan prediction and the Self-Critique prediction, one can use different  ii) number of few-shot examples iii) the context length and iv) number of samples. Finally multiple base model can be used (this is ablated in \ref{XXXX}).

In this subsection, we summarize the experimental setup. We leave an ablation study on the prompt design for the final discussion of the results in Section \ref{sec:discussion}. 
We determined the number of few-shot exemplars based on the optimal performance observed in Blocksworld, as depicted in Figure \ref{fig:num-shots-examples}, where peak performance was achieved with approximately 16 exemplars. For Blocksworld and Mini-grid-easy, performance improves with an increased number of shots but begins to plateau early, with a noticeable leveling off at just 8 shots, and even shows a slight decline thereafter with 32-shots. 
% \hanie{discussion/thoughts on why plateau happens would be useful} +1 
The Self-Critique iteration count was limited to 10 to constrain execution time, given that additional iterations produced only smaller gains in performance. Fig.~\ref{fig:stepwise-accuracy} suggests that the potential of our method can still be leveraged with more improvement steps. This count was consistently applied across other datasets to enhance the cost efficiency of the experiments, particularly in terms of token usage. For datasets such as Minigrid and Logistics, however, more shots could have been beneficial.
For the Self-Critique prompt, we explored in this work 0-shot and 8-shots on Blocksworld  and investigated multiple prompt templates, ablated in Section \ref{sec:discussion}. We ultimately selected zero-shot Self-Critique prompts because this approach does not require bootstrapping with few-shot prompts for all tasks, making it simpler and more directly applicable to new planning tasks.

% \pierre{do we have any experiment on this?}: BB: no systematic results

% We consider the number of prompts as the more serious factor impacting length of latency (due to requiring multiple calls), and further, the number of tokens is multiplied by calls to the LLM. 

% In our experiments, we explore the performance impact of increasing the number of exemplars. Figure~\ref{fig:bw-num-shots} presents graphs depicting the base method performance using Gemini 1.5 Pro and Gemma-2 27b as the LLMs.

% In experiments involving an oracle-critique (referenced in Figure~\ref{fig:bw-num-shots} under Gemini 1.5 Pro Validator), we utilize an external validator. Feedback is provided to the prompt in subsequent rounds by stating whether the goal has been reached and thus the plan is correct, the goal has not been reached but the plan could still be executed, or if the plan execution fails.

\subsection{Ablation Study}
\label{sec:discussion}

In previous section, the focus was on experiments to determine the optimal performance and access the performance of the proposed method.  In the following, we provide a discussion on the elements critical to the performance of the Self-Critique method, and summarize the results.
We demonstrate several ablation studies on prompt design and characterize performance in terms of required LLM calls. We investigate the statistical behaviour of the Self-Critique method over the course of the prediction and Self-Critique steps.

\begin{figure}[t]
    \centering
    \includegraphics[width=\textwidth]{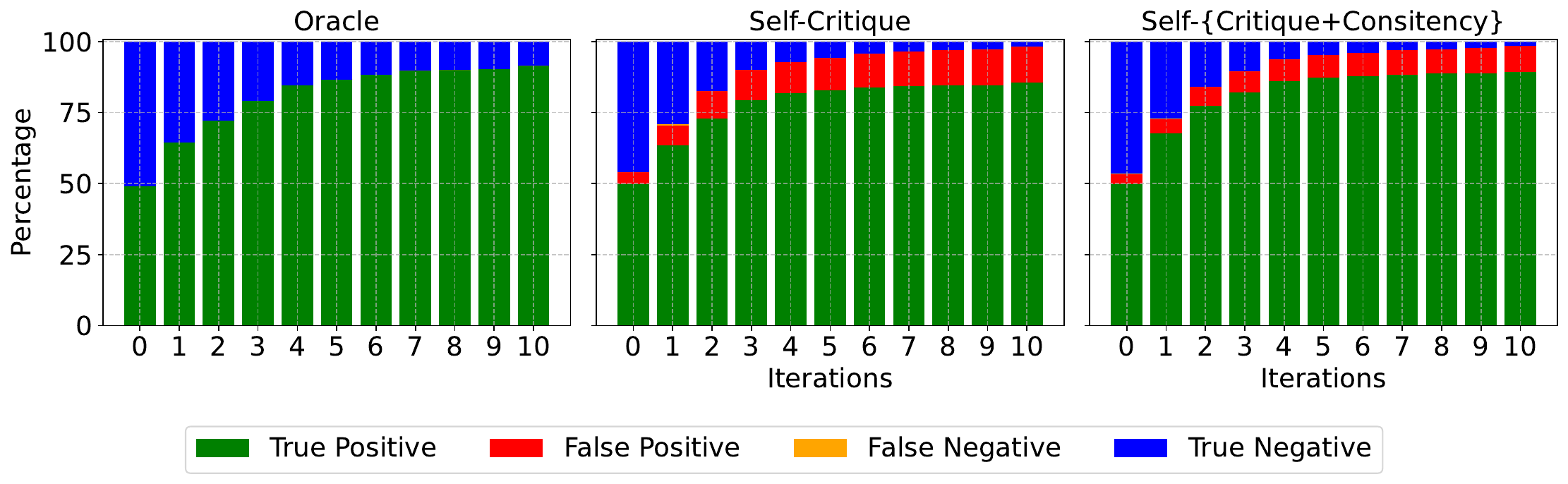}
    \caption{
    Accuracy of the Self-Critique process over the course of 10 Self-Critique steps. For these ablations, we use the Blocksworld validation set with 3-7 blocks, a 16-shot plan prediction prompt, and an 8-shot Self-Critique prompt. Overall, we observe relatively high accuracy and recall, but lower precision. Figure \ref{fig:error_analysis_full} in Appendix~\ref{ap:ablation:full} shows this error analysis on more ablations in terms of domain definition, temperature, instructions, number of steps, etc.} 
    \label{fig:error_analysis_short}
\end{figure}

Table~\ref{tab:critique_prompt_ablation} summarizes the results of the ablation study, in which we ablate various elements of our
Self-Critique method to investigate their contributions and potential sources of gains. We conduct the
test on 1000 problems from the validation set and provided 95\% Confidence Intervals.

%Figure~\ref{fig:error_analysis_short} summarizes the results of the ablation study on self critique, for more detailed ablation on various elements of the method and their effect on true positive, false positive, false negative and true negative see Figure~\ref{fig:error_analysis_full} in Appendix~\ref{ap:ablation:full}. 

\begin{table}
\caption{Ablation study on the validation set for Blocksworld 3-7. We also report the accuracy of the first step without self-critique to account for non-deterministic predictions. Unless stated otherwise, we use a temperature of 0.0 for the LLM calls.  A low temperature is expected to make the model more deterministic, but less diverse in its predictions. The highest accuracy was achieved using a self-critique method that employs self-consistency in each critique, where outcomes are determined by votes. }
\label{tab:critique_prompt_ablation}
\begin{center}
\begin{tabular}{llrrr}

  \multicolumn{1}{c}{\bf Ablation (prompts in Appendix)} &   \multicolumn{1}{c}{\bf 1st step} & \multicolumn{1}{c}{\bf 11th step} & \multicolumn{1}{c}{\bf LLM calls} \\

\hline 
8-shot self-critic with self-consistency 5 votes (\ref{app:self-eval-prompt-few-shot}) &  57.1$\pm$3.1 &    84.6$\pm$2.2 &14.0k\\ % 14015
8-shot self-critic with self-consistency 2 votes (\ref{app:self-eval-prompt-few-shot}) &  56.1$\pm$3.1 &    84.5$\pm$2.2 & 12.2k\\ % 12294
{\bf no} self-consistency, 8-shot self-critic (\ref{app:self-eval-prompt-few-shot}) & 56.2$\pm$3.1 &  79.7$\pm$2.5& 6.3k \\ %6295
\hline
{\bf no} few-shot self-critic $\rightarrow$ 0-shot self-critic (\ref{app:self-eval-prompt-no-few-shot})    & 55.7$\pm$3.1 & 79.5$\pm$2.5 & 6.1k \\ % 6128
{\bf temperature 0.2}; 0-shot self-critic (\ref{app:self-eval-prompt-no-few-shot})    & 56.1$\pm$3.1 & 76.1$\pm$2.7 & 6.4k \\ % 6428
{\bf temperature 0.5}; 0-shot self-critic (\ref{app:self-eval-prompt-no-few-shot})    & 56.9$\pm$3.1 & 78.2$\pm$2.7 & 6.1k \\ % 6428
\hline
{\bf no} Domain Definition (\ref{app:self-eval-prompt-0-shot-no-domain-definition})      & 56.0$\pm$3.1 & 74.4$\pm$2.7 & 5.8k\\ %5808
{\bf no} 3-Step Instruction  (\ref{app:self-eval-prompt-ablation-no-dd-no-3step})  & 55.2$\pm$3.1 & 64.0$\pm$3.1 & 4.1k \\ %4096
 {\bf no} 'verify each action' (\ref{app:self-eval-prompt-ablation-very-plan}) & 56.1$\pm$3.1 & 57.5$\pm$3.1 &  3.4k \\ %3428
\end{tabular}
\end{center}
\end{table}

{\bf Self-Critique evaluation.}  Figure~\ref{fig:error_analysis_short} displays the accuracy of the plan and the Self-Critique method, over the course of 10 self-improvement iterations. 
The accumulation of false positives visualizes the most frequent error in the approach, whereas the Oracle represents the ideal case, showing the upper bound of accuracy achievable using Self-Critique. False negatives occur very rarely.

%Note the accumulation of false positives for Self-Critique approaches compared to the model that employs the oracle for feedback, demonstrating the effect of Self-Critique for self-improvement. %Dilan: rephrased the following, feel free to revert back.
%displays the evolution of the Self-Critique accuracy (compared to the ground-truth critique provided by the oracle) over the course of 11 iterations. The graphs offer insights into the development of correct (True) and incorrect (False) predictions.

{\bf Self-consistency.} We also study self-consistency \citep{wang2023selfconsistency_cot} in the Self-Critique process, utilizing voting to classify outcomes as either {\em wrong/goal not reached} or {\em correct}. In the event of a tie, the outcome is designated as {\em wrong}, which initiates a new prompt/critique step.
Figures \ref{fig:stepwise-accuracy} and \ref{fig:error_analysis_short}  and Table \ref{tab:critique_prompt_ablation} show the impact of adding self-consistency to the Self-Critique, providing an additional improvement of nearly 5\% absolute.

{\bf Foundational Models.} To ascertain the efficacy of the proposed Self-Critique method, we run experiments across different foundational models. We demonstrate substantial improvements  across the SoTA foundational models Gemini 1.5, GPT-4o and Claude 3.5 Sonnet.  
%The details are included in Appendix \ref{ap:other-models}.
Table \ref{tab:llm-comparison} shows the results for GPT-4o and Claude 3.5 Sonnet. Appendix~\ref{ap:gemma_results} includes the results for experiments on Gemma 2-27b.
% \subsection{Results on further LLMs'}
% \label{ap:other-models}

\begin{table}[h]
\caption{The performance of various LLMs to confirm the applicability of our method across different platforms. The focus is primarily on validating the method rather than comparing the performance across LLMs. We use 16 few-shot for the plan prediction and the 0-shot critic (\ref{app:self-eval-prompt-no-few-shot}) and Blocksworld 3-5 \cite{valmeekam2023planning} as dataset.  }
\label{tab:llm-comparison}
\begin{center}
\setlength{\tabcolsep}{4pt} % default is 6pt
\begin{tabular}{lcccc}
\multicolumn{1}{c}{} & \multicolumn{2}{c}{\bf GPT-4o }& \multicolumn{2}{c}{\bf Claude 3.5 Sonnet} \\
\cmidrule(lr){2-3} \cmidrule(lr){4-5}

% {\bf Method} & \multicolumn{1}{c}{\bf few-shot} & {\bf few-shot}  \\

% \hline
{\bf Dataset} & {\bf No-Critique} & {\bf Self-Critique} & {\bf No-Critique} & {\bf Self-Critique}  \\ 
% \hline \\
\hline 
Blocksworld 3-5 &  42.8$\pm$3.9 & 64.2$\pm$3.8  &68.0$\pm$3.7 & 89.5 $\pm$2.5   

% Mystery BW & 4 & 0 & 8  &22.3$\pm$3.3 & 35.2$\pm$3.8 & 37.8$\pm$3.9 &37.3$\pm$3.9\\ % 
\end{tabular}
\end{center}
\end{table}

{\bf Complexity.}  Equation (\ref{eq:Self-Critique}) defines the upper limit of LLM calls $q$, as the measure of complexity for the self-improvement process. The actual calls to the LLM, documented in Table~\ref{tab:critique_prompt_ablation}, are significantly lower due to the process being stopped once the LLM deems a solution correct.
The upper limit of calls to the LLM per iteration is linearly related to twice the maximum number of allowed steps consisting of a prediction and a subsequent Self-Critique. 
Note, for practical reasons, we stop the iterations after $k$ steps (10 in this study) to reduce the required context length and runtime.
%\footnote{In scenarios where the process is halted early after reaching this predefined limit $k$, one call should be subtracted for the Self-Critique from the total query count $q$, which is not accounted for in Equation (\ref{eq:Self-Critique}).}
\begin{equation}
\label{eq:Self-Critique}
\text{q} = 2 \times s 
\end{equation}

When self-consistency is employed, it introduces an additional multiplicative constant factor $c$ for the number of self-consistency calls as shown in Equation (\ref{eq:Self-Critique-self-consistency}). It is important to note that self-consistency does not increase latency, as these operations can be executed in parallel.
\begin{equation}
\label{eq:Self-Critique-self-consistency}
\text{q} = s + c \times s. 
\end{equation}

\section{Conclusion}
\label{sec:conclusion}

This work shows that self-improvement using intrinsic self-critique can significantly enhances performance on standard planning benchmarks, when properly implemented. We obtain substantial performance gains on all of the studied benchmarks. Problems that had previously been challenging are now solvable with a high accuracy presenting SoTA for the model class considered, such as in Blocksworld 3-5, where we achieve an 89.3\% success rate employing self-critique with self-consistency. 
This work is also the first to demonstrate that LLMs can solve Mystery Blocksworld problems with 22\% accuracy and achieve substantial accuracy gains, reaching 37.8\%, when self-improvement using self-critique and self-consistency is employed.

Furthermore, there is potential to swap out in-context learning for more sophisticated planning methods, such as Chain-of-Thought, Self-Consistency, or even to integrate these with advanced search-based algorithms like Monte-Carlo Tree Search to further improve accuracy or tackle even more complex tasks \citep{ Coulom2007,wei2022chain,madaan2023selfrefine}.

%Furthermore, there is potential to incorporate more sophisticated LLM-based planning methods, such as Chain-of-Thought and Self-Consistency, or even to integrate with advanced search-based algorithms like Monte-Carlo Tree Search.

Lastly, this work not only demonstrates the viability of self-improvement using self-critique in enhancing planning accuracy but also lays the groundwork for a new paradigm in AI planning. By bridging the gap between symbolic planning and language models, we open up possibilities for tackling more complex, real-world planning scenarios and pushing the boundaries of AI problem-solving capabilities by leveraging LLMs.

% \begin{enumerate}
%     \item This work pioneered to enhance planning accuracy via self-critique. The accuracy gain are substantial. 
%     \item For planning, the method leverages the formal descriptions to verification in tree step approach. 
%     \item The method benefits propbably also from foundation model knowlege of pddl widely use in AI since the 1980 and hence probably in the training data. 
%     \item Now, it becomes possible to tackle more difficult planning problems for which planning domains allows to scale up the problem difficult problems for instance in Blocksworld by increasing the number of blocks and plan length or logistics, the number of places, packages, etc.  \item We believe that this approach not only strengthens the foundation for incremental enhancements but might also synergize well with other strong planning methods like Chain-of-Thought and Self-Consistency, a hypothesis left for future work.
% \end{enumerate}

\newpage
\bibliography{iclr2025_conference, references}
\newpage
\appendix
\section{Implementation details}
\label{sec:implementation_details}

\subsection{Prompts}

\begin{examplebox}[app:self-eval-prompt-few-shot]{Self-Critique Prompt with few-shot exemplars}

% \begin{smallverbatim}
Given the domain definition:\\
\{domain\_pddl\}\\

For instance, we verify the following steps.\\
\{self\_evaluations\_exemplars\}\\
\\
So, for each action:\\
1. Take the action and its preconditions from the domain definition for the specific action.\\
2. Verify whether the preconditions are met for the action.\\
3. Apply the action and provide the resulting state.\\

The problem to solve:\\
\{instance\} \\

The suggested solution:\\
\{plan\}\\

Please carefully evaluate the plan. Verify each step as described above. Do not stop until each action is verified; please *do not* omit steps.
Conclude with the assessment literally either with 'the plan is correct', 'the plan is wrong', or 'goal not reached'.
% \end{smallverbatim}
\end{examplebox}

\begin{examplebox}[app:self-eval-prompt-no-few-shot]{Self-Critique 0-shot Prompt (without exemplars) with Domain Definition}

% \begin{smallverbatim}
Given the domain definition:\\
\{domain\_pddl\}\\
\\
So, for each action:\\
1. Take the action and its preconditions from the domain definition for the specific action.\\
2. Verify whether the preconditions are met for the action.\\
3. Apply the action and provide the resulting state.\\

The problem to solve:\\
\{instance\} \\

The suggested solution:\\
\{plan\}\\

Please carefully evaluate the plan. Verify each step as described above. Do not stop until each action is verified; please *do not* omit steps.
Conclude with the assessment literally either with 'the plan is correct', 'the plan is wrong', or 'goal not reached'.
% \end{smallverbatim}
\end{examplebox}

\begin{examplebox}[app:self-eval-prompt-0-shot-no-domain-definition]{Self-Critique 0-shot Prompt without Domain Definition }

So, for each action:\\
1. Take the action and its preconditions from the domain definition for the specific action.\\
2. Verify whether the preconditions are met for the action.\\
3. Apply the action and provide the resulting state.\\

The problem to solve:\\
\{instance\} \\

The suggested solution:\\
\{plan\}\\

Please carefully evaluate the plan. Verify each step as described above. Do not stop until each action is verified; please *do not* omit steps.
Conclude with the assessment literally either with 'the plan is correct', 'the plan is wrong', or 'goal not reached'.
\end{examplebox}

\begin{examplebox}[app:self-eval-prompt-ablation-no-dd-no-3step]{Self-Critique Prompt: Ablation of Domain Definition and numbered Instructions}

The problem to solve:\\
\{instance\} \\

The suggested solution:\\
\{plan\}\\

Please carefully evaluate the plan. Verify each action. Do not stop until each action is verified; please *do not* omit steps.
Conclude with the assessment literally either with 'the plan is correct', 'the plan is wrong', or 'goal not reached'.

\end{examplebox}

\begin{examplebox}[app:self-eval-prompt-ablation-very-plan]{Self-Critique Prompt: Ablation of Domain Definition and numbered Instructions}

The problem to solve:\\
\{instance\} \\

The suggested solution:\\
\{plan\}\\

Please carefully evaluate the plan. Verify the plan as chain-of-thought. 
Conclude with the assessment literally either with 'the plan is correct', 'the plan is wrong', or 'goal not reached'.

\end{examplebox}

\section{Few-shot prompts}
\label{app:few-shot-example-prompts}
\begin{examplebox}[app:few-shot-example-bw]{Few-shot Planning Prompt Example}
The domain definition:\\
(define (domain blocksworld-4ops)\\
  (:requirements :strips)\\
  (:predicates (clear ?x) (ontable ?x) (handempty) (holding ?x)
               (on ?x ?y))\\

  (:action pick-up\\
    :parameters (?ob)\\
    :precondition (and (clear ?ob) (ontable ?ob) (handempty))\\
    :effect (and (holding ?ob) (not (clear ?ob)) (not (ontable ?ob))
                 (not (handempty))))\\

  (:action put-down\\
    :parameters  (?ob)\\
    :precondition (holding ?ob)\\
    :effect (and (clear ?ob) (handempty) (ontable ?ob)
                 (not (holding ?ob))))\\

  (:action stack\\
    :parameters  (?ob ?underob)\\
    :precondition (and (clear ?underob) (holding ?ob))\\
    :effect (and (handempty) (clear ?ob) (on ?ob ?underob)
                 (not (clear ?underob)) (not (holding ?ob))))\\

  (:action unstack\\
    :parameters  (?ob ?underob)\\
    :precondition (and (on ?ob ?underob) (clear ?ob) (handempty))\\
    :effect (and (holding ?ob) (clear ?underob)
                 (not (on ?ob ?underob)) (not (clear ?ob)) (not (handempty))))\\

Example of a problem and its solution (plan):\\
(define (problem BW-rand-6)\\
  (:domain blocksworld-4ops)\\
  (:objects b1 b6 b4 b3 b5 b2)\\
  (:init\\
    (clear b5)\\
    (clear b2)\\
    (clear b3)\\
    (ontable b1)\\
    (on b6 b1)\\
    (on b3 b4)\\
    (ontable b2)\\
    (ontable b5)\\
    (on b4 b6)\\
    (handempty)\\
  )\\
  (:goal (and\\
    (on b3 b2)\\
    (on b1 b3)\\
    (on b6 b1)\\
    (on b4 b5)\\
  ))\\
)\\
The plan without formatting:\\
(unstack b3 b4)\\
(stack b3 b2)\\
(unstack b4 b6)\\
(stack b4 b5)\\
(unstack b6 b1)\\
(put-down b6)\\
(pick-up b1)\\
(stack b1 b3)\\
(pick-up b6)\\
(stack b6 b1)\\

... ; more few-shots\\

And now the problem for you to solve. Please solve the following problem:\\
(define (problem BW-rand-6)\\
(:domain blocksworld-4ops)\\
(:objects b5 b1 b4 b2 b3 b6)\\
(:init\\
(on b4 b1)\\
(handempty)\\
(ontable b6)\\
(on b2 b4)\\
(clear b3)\\
(ontable b5)\\
(on b3 b2)\\
(clear b6)\\
(on b1 b5)\\
)\\
(:goal (and\\
(on b4 b2)\\
(on b1 b4)\\
(on b5 b1)\\
(on b3 b5)\\
))\\

Provide only the plan. Do not add numbers, items or any additional text. Your plan as plain text without formatting:

\end{examplebox}

\subsection{LLM details}
\label{ap:llm_details}

If a prompt exceeds this length limit at any $step$, we prematurely terminate the process and conduct the final assessment during that iteration, which may result in a suboptimal solution and reduced accuracy. This limitation impacts the datasets Mini-grid, due to the extensive length of its problem definitions, but it does not affect Blocksworld when limited to 10 iterations.

\section{Additional Results}

\subsection{Gemma results} 

\label{ap:gemma_results}
\begin{figure}[ht]
     \centering
    \begin{subfigure}[b]{0.495\textwidth}
        \centering
        \includegraphics[width=\textwidth]{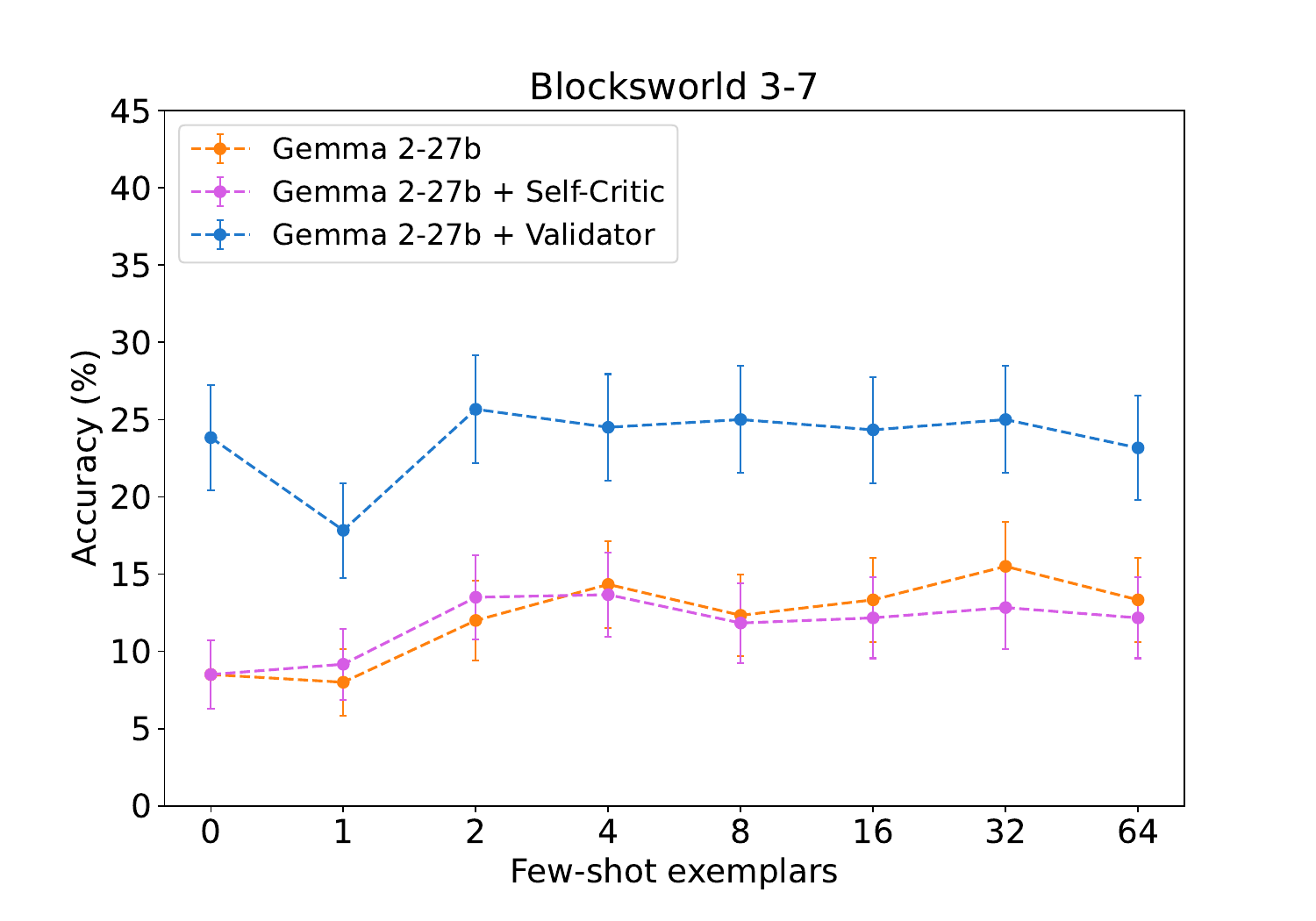}
        % \caption{\centering Planning results using Blocksworld.}  
        \label{fig:gemma-bw-3-7}  
    \end{subfigure}
     \centering
    \begin{subfigure}[b]{0.495\textwidth}
        \centering
        \includegraphics[width=\textwidth]{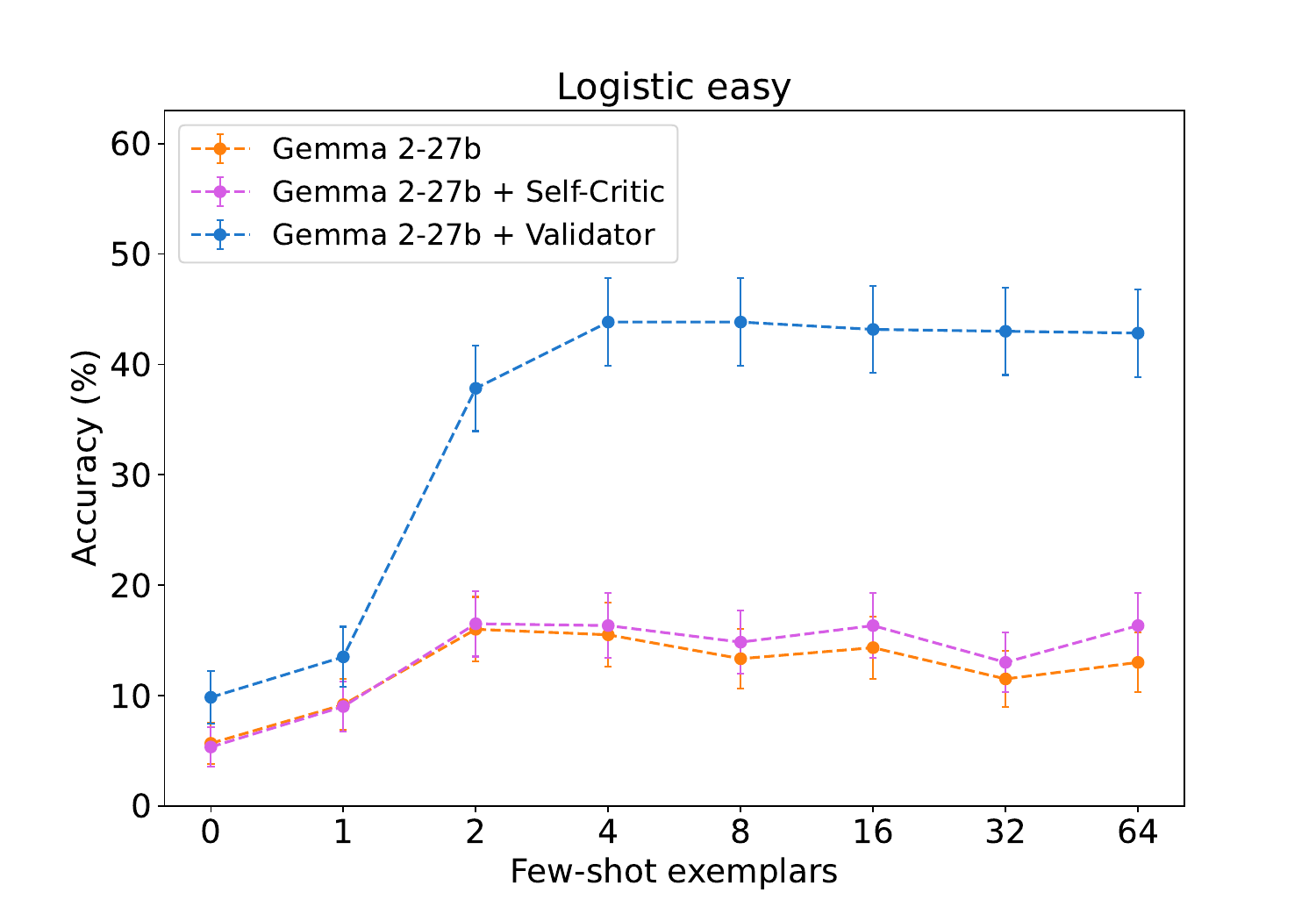}
        % \caption{\centering Planning results using Blocksworld.}  
        \label{fig:gemma-lg4-logistics}  
    \end{subfigure}
    \caption{Exploration of planning performance using Gemma 2-27b. 
    }
    \label{fig:bw-num-shots}    
\end{figure}

\subsection{Scaling the context length}
We conduct ablation studies on the context length of the model. Given that these experiments are resource-intensive, we limit them to the Minigrid environment.
\begin{figure}[ht]
    \centering
    \includegraphics[width=0.6\textwidth]{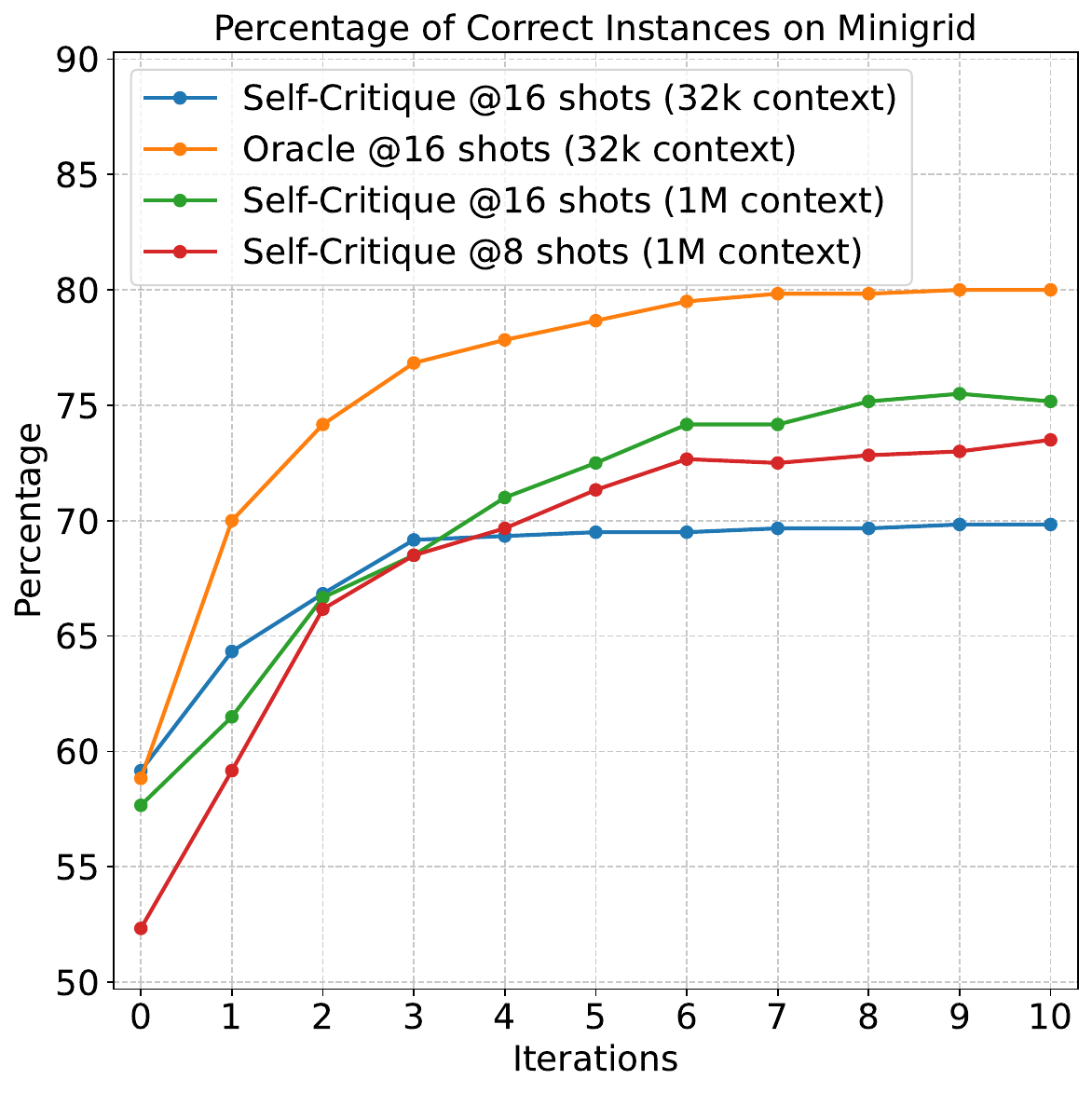}
    \caption{
Performance on Minigrid using 8-shot and 16-shot examples for the planning prompts, with a context length limit of 32k vs 1 million tokens, plotted across self-improvement iterations. } 
    \label{fig:ablation_minigrid_context_length}
\end{figure}

\subsection{Natural Language and PDDL} 

Table \ref{tab:natural_language_pddl}
compares results for Gemini 1.5 Pro using the prompts of \cite{valmeekam2023planning} in Natural Language as well as PDDL.
The performance of the Natural Language version is lower than that of the PDDL version when employing a 1-shot approach. The table also confirms the efficacy of the self-critique method in Natural Language applications, as demonstrated by an improvement in accuracy from 20.3 to 29.7.

\begin{table}
\centering
\caption{Natural Language and PDDL using Blocksworld 3-5. We use the 1-shot prompts for natural language \cite{valmeekam2023planning} and the self-critique templates from \cite{stechly2024selfverification} for this study. The presented results on PDDL also employ 1-shot prompts and templates from \cite{stechly2024selfverification}. Natural Language and PDDL using Blocksworld 3-5. }
\label{tab:natural_language_pddl}
\begin{tabular}{llrr}

\textbf{Dataset} & \textbf{Prompt} &\textbf{No-Critique} & \textbf{Self-Critic} \\
\hline
Natural Language & Formatted & 18.5$\pm$3.1 &19.2$\pm$3.1 \\
Natural Language & CoT+Formatted & 20.3$\pm$3.2 & 29.7$\pm$3.7 \\
\hline
PDDL             & Formatted & 40.3$\pm$3.9 & 47.3$\pm$4.0\\
PDDL             & CoT+Formatted & 39.2$\pm$4.0 & 65.0$\pm$3.8 \\

\hline
\end{tabular}
\end{table}

\subsection{Error analysis (full)} \label{ap:ablation:full}
\begin{figure}[ht]
    \centering
    \includegraphics[width=\textwidth]{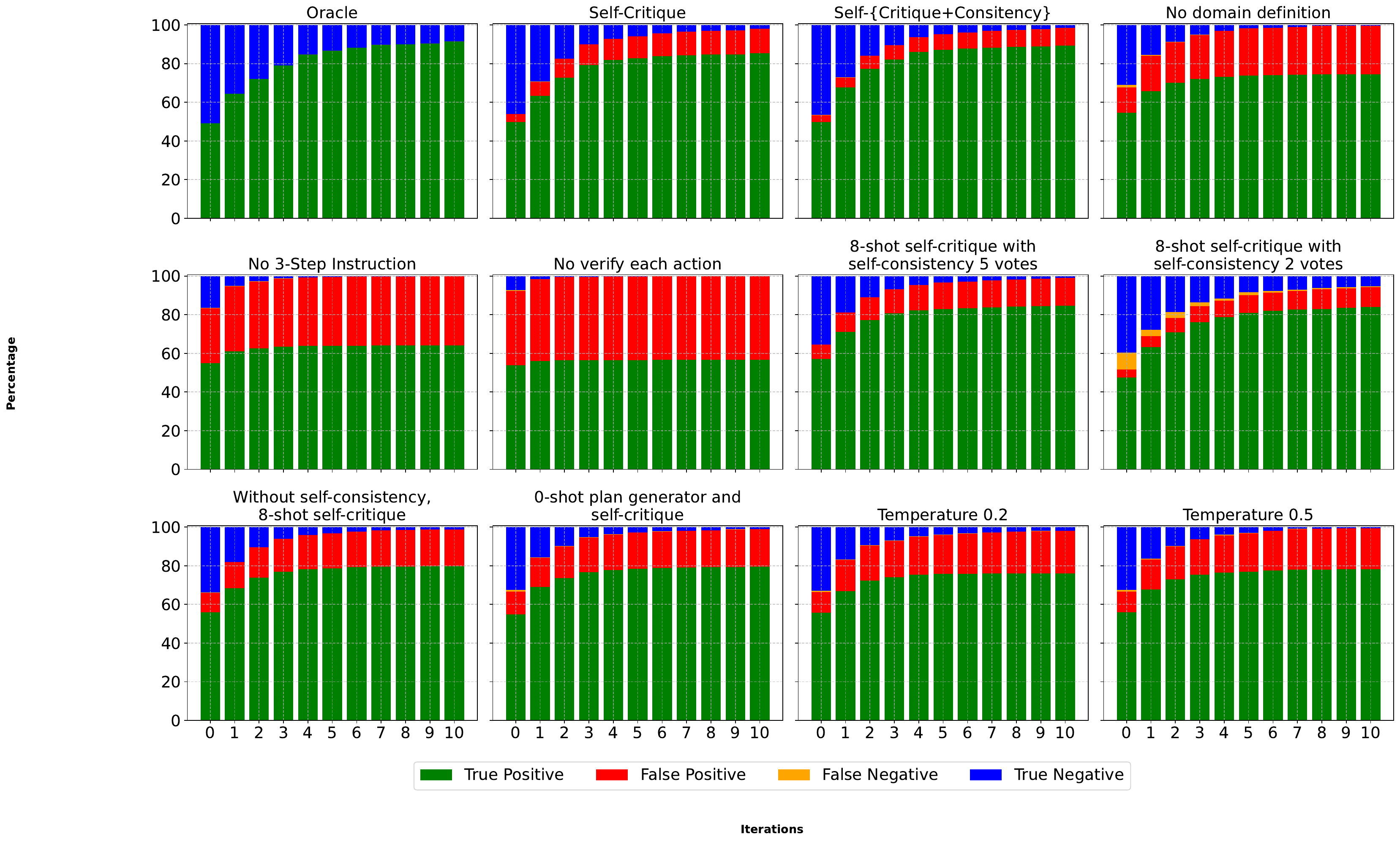}
    \caption{
    Analysis of the Self-Critique over 11 rounds of improvement steps. For these exploitative experiments, we used the Blocksworld validation set with 3-7 blocks, a 16-shot plan prediction prompt, and an 8-shot self-critique prompt. Overall, we observe relatively high accuracy and recall, but lower precision.} 
    \label{fig:error_analysis_full}
\end{figure}

\section{Example of self-improvement trace Blocksworld}

\begin{examplebox}[app:sample-trace]{Self-Improvment Trace}
\footnotesize
\begin{verbatim}
    
The domain definition:
  (define (domain blocksworld-4ops)
  (:requirements :strips)
(:predicates (clear ?x)
             (ontable ?x)
             (handempty)
             (holding ?x)
             (on ?x ?y))

(:action pick-up
  :parameters (?ob)
  :precondition (and (clear ?ob) (ontable ?ob) (handempty))
  :effect (and (holding ?ob) (not (clear ?ob)) (not (ontable ?ob))
               (not (handempty))))

(:action put-down
  :parameters  (?ob)
  :precondition (holding ?ob)
  :effect (and (clear ?ob) (handempty) (ontable ?ob)
               (not (holding ?ob))))

(:action stack
  :parameters  (?ob ?underob)
  :precondition (and (clear ?underob) (holding ?ob))
  :effect (and (handempty) (clear ?ob) (on ?ob ?underob)
               (not (clear ?underob)) (not (holding ?ob))))

(:action unstack
  :parameters  (?ob ?underob)
  :precondition (and (on ?ob ?underob) (clear ?ob) (handempty))
  :effect (and (holding ?ob) (clear ?underob)
               (not (on ?ob ?underob)) 
               (not (clear ?ob)) (not (handempty)))))

Example of a problem and its solution (plan):
(define (problem BW-rand-7)
(:domain blocksworld-4ops)
(:objects b6 b3 b1 b7 b4 b2 b5)
(:init
(on b7 b1)
(ontable b6)
(clear b3)
(ontable b4)
(on b5 b2)
(clear b6)
(clear b7)
(ontable b3)
(handempty)
(clear b5)
(on b2 b4)
(ontable b1)
)
(:goal (and
(on b3 b2)
(on b5 b1)
))
)

The plan without formatting:
(unstack b7 b1)
(put-down b7)
(unstack b5 b2)
(stack b5 b1)
(pick-up b3)
(stack b3 b2)


Example of a problem and its solution (plan):
(define (problem BW-rand-5)
(:domain blocksworld-4ops)
(:objects b4 b1 b2 b5 b3)
(:init
(handempty)
(on b2 b1)
(clear b2)
(ontable b4)
(clear b3)
(on b3 b5)
(on b1 b4)
(ontable b5)
)
(:goal (and
(on b2 b1)
(on b5 b4)
(on b3 b5)
))
)

The plan without formatting:
(unstack b2 b1)
(put-down b2)
(unstack b1 b4)
(put-down b1)
(pick-up b2)
(stack b2 b1)
(unstack b3 b5)
(put-down b3)
(pick-up b5)
(stack b5 b4)
(pick-up b3)
(stack b3 b5)

;
; further few-shots omitted.
; ...

And now the problem for you to solve. Please solve the 
following problem:
(define (problem BW-rand-5)
(:domain blocksworld-4ops)
(:objects b4 b1 b2 b5 b3)
(:init
(on b1 b4)
(handempty)
(ontable b4)
(on b2 b1)
(on b5 b2)
(ontable b3)
(clear b3)
(clear b5)
)
(:goal (and
(on b2 b5)
(on b3 b2)
(on b1 b4)
))
)

Provide only the plan. Do not add numbers, items or any additional
text. 
Your plan as plain text without formatting:
The clean plan:
(unstack b5 b2)
(put-down b5)
(unstack b2 b1)
(put-down b2)
(pick-up b3)
(stack b3 b2)
(pick-up b5)
(stack b5 b3)
(pick-up b2)
(stack b2 b5)
(pick-up b1)
(stack b1 b4)
## plan evaluation:

**step 1: (unstack b5 b2)**

1. **action and preconditions:**
   - action: unstack ?ob ?underob
   - preconditions: (and (on ?ob ?underob) (clear ?ob) (handempty))

2. **verification:**
   - (on b5 b2): **true** (from initial state)
   - (clear b5): **true** (from initial state)
   - (handempty): **true** (from initial state)
   - **all preconditions are met.**

3. **resulting state:**
   - (holding b5)
   - (clear b2)
   - (not (on b5 b2))
   - (not (clear b5))
   - (not (handempty))
   - **other facts from the initial state remain unchanged.**

**step 2: (put-down b5)**

1. **action and preconditions:**
   - action: put-down ?ob
   - preconditions: (holding ?ob)

2. **verification:**
   - (holding b5): **true** (from the previous step)
   - **precondition is met.**

3. **resulting state:**
   - (clear b5)
   - (handempty)
   - (ontable b5)
   - (not (holding b5))
   - **other facts from the previous step remain unchanged.**

**step 3: (unstack b2 b1)**

1. **action and preconditions:**
   - action: unstack ?ob ?underob
   - preconditions: (and (on ?ob ?underob) (clear ?ob) (handempty))

2. **verification:**
   - (on b2 b1): **true** (from initial state)
   - (clear b2): **true** (from the previous step)
   - (handempty): **true** (from the previous step)
   - **all preconditions are met.**

3. **resulting state:**
   - (holding b2)
   - (clear b1)
   - (not (on b2 b1))
   - (not (clear b2))
   - (not (handempty))
   - **other facts from the previous step remain unchanged.**

**step 4: (put-down b2)**

1. **action and preconditions:**
   - action: put-down ?ob
   - preconditions: (holding ?ob)

2. **verification:**
   - (holding b2): **true** (from the previous step)
   - **precondition is met.**

3. **resulting state:**
   - (clear b2)
   - (handempty)
   - (ontable b2)
   - (not (holding b2))
   - **other facts from the previous step remain unchanged.**

**step 5: (pick-up b3)**

1. **action and preconditions:**
   - action: pick-up ?ob
   - preconditions: (and (clear ?ob) (ontable ?ob) (handempty))

2. **verification:**
   - (clear b3): **true** (from initial state)
   - (ontable b3): **true** (from initial state)
   - (handempty): **true** (from the previous step)
   - **all preconditions are met.**

3. **resulting state:**
   - (holding b3)
   - (not (clear b3))
   - (not (ontable b3))
   - (not (handempty))
   - **other facts from the previous step remain unchanged.**

**step 6: (stack b3 b2)**

1. **action and preconditions:**
   - action: stack ?ob ?underob
   - preconditions: (and (clear ?underob) (holding ?ob))

2. **verification:**
   - (clear b2): **true** (from the previous step)
   - (holding b3): **true** (from the previous step)
   - **all preconditions are met.**

3. **resulting state:**
   - (handempty)
   - (clear b3)
   - (on b3 b2)
   - (not (clear b2))
   - (not (holding b3))
   - **other facts from the previous step remain unchanged.**

**step 7: (pick-up b5)**

1. **action and preconditions:**
   - action: pick-up ?ob
   - preconditions: (and (clear ?ob) (ontable ?ob) (handempty))

2. **verification:**
   - (clear b5): **true** (from step 2)
   - (ontable b5): **true** (from step 2)
   - (handempty): **true** (from the previous step)
   - **all preconditions are met.**

3. **resulting state:**
   - (holding b5)
   - (not (clear b5))
   - (not (ontable b5))
   - (not (handempty))
   - **other facts from the previous step remain unchanged.**

**step 8: (stack b5 b3)**

1. **action and preconditions:**
   - action: stack ?ob ?underob
   - preconditions: (and (clear ?underob) (holding ?ob))

2. **verification:**
   - (clear b3): **true** (from step 6)
   - (holding b5): **true** (from the previous step)
   - **all preconditions are met.**

3. **resulting state:**
   - (handempty)
   - (clear b5)
   - (on b5 b3)
   - (not (clear b3))
   - (not (holding b5))
   - **other facts from the previous step remain unchanged.**

**step 9: (pick-up b2)**

1. **action and preconditions:**
   - action: pick-up ?ob
   - preconditions: (and (clear ?ob) (ontable ?ob) (handempty))

2. **verification:**
   - (clear b2): **false** (on b3 b2 from step 6)
   - (ontable b2): **false** (on b3 b2 from step 6)
   - (handempty): **true** (from the previous step)
   - **preconditions are not met.**

**the plan is wrong** because the preconditions for 
picking up b2 in step 9 are not met. 

Please can you explain the error and fix it.
The clean plan:
(unstack b5 b2)
(put-down b5)
(unstack b2 b1)
(put-down b2)
(pick-up b3)
(stack b3 b2)
(pick-up b5)
(stack b5 b3)
(unstack b3 b2)
(put-down b3)
(pick-up b2)
(stack b2 b5)
(pick-up b1)
(stack b1 b4)
## plan verification:

**step 1: (unstack b5 b2)**

1. **action and preconditions:** 
   - action: unstack ?ob ?underob
   - preconditions: (and (on ?ob ?underob) (clear ?ob) (handempty))
2. **verification:**
   - (on b5 b2): **true** (from initial state)
   - (clear b5): **true** (from initial state)
   - (handempty): **true** (from initial state)
   - **preconditions met**
3. **resulting state:**
   - (on b1 b4)
   - (ontable b4)
   - (on b2 b1)
   - (ontable b3)
   - (clear b3)
   - **(holding b5)**
   - **(clear b2)**
   - **(not (on b5 b2))**
   - **(not (clear b5))**
   - **(not (handempty))**

**step 2: (put-down b5)**

1. **action and preconditions:**
   - action: put-down ?ob
   - preconditions: (holding ?ob)
2. **verification:**
   - (holding b5): **true** (from previous step)
   - **preconditions met**
3. **resulting state:**
   - (on b1 b4)
   - (ontable b4)
   - (on b2 b1)
   - (ontable b3)
   - (clear b3)
   - **(ontable b5)**
   - (clear b2)
   - **(clear b5)**
   - **(handempty)**

**step 3: (unstack b2 b1)**

1. **action and preconditions:**
   - action: unstack ?ob ?underob
   - preconditions: (and (on ?ob ?underob) (clear ?ob) (handempty))
2. **verification:**
   - (on b2 b1): **true** (from previous step)
   - (clear b2): **true** (from previous step)
   - (handempty): **true** (from previous step)
   - **preconditions met**
3. **resulting state:**
   - (on b1 b4)
   - (ontable b4)
   - (ontable b3)
   - (ontable b5)
   - (clear b3)
   - (clear b5)
   - **(holding b2)**
   - **(clear b1)**
   - **(not (on b2 b1))**
   - **(not (clear b2))**
   - **(not (handempty))**

**step 4: (put-down b2)**

1. **action and preconditions:**
   - action: put-down ?ob
   - preconditions: (holding ?ob)
2. **verification:**
   - (holding b2): **true** (from previous step)
   - **preconditions met**
3. **resulting state:**
   - (on b1 b4)
   - (ontable b4)
   - (ontable b3)
   - (ontable b5)
   - (clear b3)
   - (clear b5)
   - **(ontable b2)**
   - (clear b1)
   - **(clear b2)**
   - **(handempty)**

**step 5: (pick-up b3)**

1. **action and preconditions:**
   - action: pick-up ?ob
   - preconditions: (and (clear ?ob) (ontable ?ob) (handempty))
2. **verification:**
   - (clear b3): **true** (from previous step)
   - (ontable b3): **true** (from previous step)
   - (handempty): **true** (from previous step)
   - **preconditions met**
3. **resulting state:**
   - (on b1 b4)
   - (ontable b4)
   - (ontable b5)
   - (ontable b2)
   - (clear b5)
   - (clear b2)
   - (clear b1)
   - **(holding b3)**
   - **(not (clear b3))**
   - **(not (ontable b3))**
   - **(not (handempty))**

**step 6: (stack b3 b2)**

1. **action and preconditions:**
   - action: stack ?ob ?underob
   - preconditions: (and (clear ?underob) (holding ?ob))
2. **verification:**
   - (clear b2): **true** (from previous step)
   - (holding b3): **true** (from previous step)
   - **preconditions met**
3. **resulting state:**
   - (on b1 b4)
   - (ontable b4)
   - (ontable b5)
   - **(on b3 b2)**
   - (clear b5)
   - **(clear b3)**
   - (clear b1)
   - **(handempty)**
   - **(not (clear b2))**

**step 7: (pick-up b5)**

1. **action and preconditions:**
   - action: pick-up ?ob
   - preconditions: (and (clear ?ob) (ontable ?ob) (handempty))
2. **verification:**
   - (clear b5): **true** (from previous step)
   - (ontable b5): **true** (from previous step)
   - (handempty): **true** (from previous step)
   - **preconditions met**
3. **resulting state:**
   - (on b1 b4)
   - (ontable b4)
   - (on b3 b2)
   - (clear b3)
   - (clear b1)
   - **(holding b5)**
   - **(not (clear b5))**
   - **(not (ontable b5))**
   - **(not (handempty))**

**step 8: (stack b5 b3)**

1. **action and preconditions:**
   - action: stack ?ob ?underob
   - preconditions: (and (clear ?underob) (holding ?ob))
2. **verification:**
   - (clear b3): **true** (from previous step)
   - (holding b5): **true** (from previous step)
   - **preconditions met**
3. **resulting state:**
   - (on b1 b4)
   - (ontable b4)
   - **(on b5 b3)**
   - (on b3 b2)
   - (clear b1)
   - **(clear b5)**
   - **(handempty)**
   - **(not (clear b3))**

**step 9: (unstack b3 b2)**

1. **action and preconditions:**
   - action: unstack ?ob ?underob
   - preconditions: (and (on ?ob ?underob) (clear ?ob) (handempty))
2. **verification:**
   - (on b3 b2): **true** (from previous step)
   - (clear b3): **false** (from previous step)
   - (handempty): **true** (from previous step)
   - **preconditions not met**

**the plan is wrong.** 

the plan fails at step 9 because the precondition 
(clear b3) is not met. block b5 is on top of b3, 
making it not clear. 

Please can you explain the error and fix it.
The clean plan:
(unstack b5 b2)
(put-down b5)
(unstack b2 b1)
(stack b2 b5)
(pick-up b3)
(stack b3 b2)
(pick-up b1)
(stack b1 b4)
## plan verification:

**step 1: (unstack b5 b2)**

1. **action and preconditions:** 
   - action: unstack ?ob ?underob
   - preconditions: (and (on ?ob ?underob) (clear ?ob) (handempty))
2. **verification:**
   - (on b5 b2): **true** (from initial state)
   - (clear b5): **true** (from initial state)
   - (handempty): **true** (from initial state)
   - **all preconditions met.**
3. **resulting state:**
   - (holding b5)
   - (clear b2)
   - (not (on b5 b2))
   - (not (clear b5))
   - (not (handempty))

**step 2: (put-down b5)**

1. **action and preconditions:**
   - action: put-down ?ob
   - preconditions: (holding ?ob)
2. **verification:**
   - (holding b5): **true** (from previous step)
   - **precondition met.**
3. **resulting state:**
   - (clear b5)
   - (handempty)
   - (ontable b5)
   - (not (holding b5))

**step 3: (unstack b2 b1)**

1. **action and preconditions:**
   - action: unstack ?ob ?underob
   - preconditions: (and (on ?ob ?underob) (clear ?ob) (handempty))
2. **verification:**
   - (on b2 b1): **true** (from initial state)
   - (clear b2): **true** (from previous step)
   - (handempty): **true** (from previous step)
   - **all preconditions met.**
3. **resulting state:**
   - (holding b2)
   - (clear b1)
   - (not (on b2 b1))
   - (not (clear b2))
   - (not (handempty))

**step 4: (stack b2 b5)**

1. **action and preconditions:**
   - action: stack ?ob ?underob
   - preconditions: (and (clear ?underob) (holding ?ob))
2. **verification:**
   - (clear b5): **true** (from step 2)
   - (holding b2): **true** (from previous step)
   - **all preconditions met.**
3. **resulting state:**
   - (handempty)
   - (clear b2)
   - (on b2 b5)
   - (not (clear b5))
   - (not (holding b2))

**step 5: (pick-up b3)**

1. **action and preconditions:**
   - action: pick-up ?ob
   - preconditions: (and (clear ?ob) (ontable ?ob) (handempty))
2. **verification:**
   - (clear b3): **true** (from initial state)
   - (ontable b3): **true** (from initial state)
   - (handempty): **true** (from previous step)
   - **all preconditions met.**
3. **resulting state:**
   - (holding b3)
   - (not (clear b3))
   - (not (ontable b3))
   - (not (handempty))

**step 6: (stack b3 b2)**

1. **action and preconditions:**
   - action: stack ?ob ?underob
   - preconditions: (and (clear ?underob) (holding ?ob))
2. **verification:**
   - (clear b2): **true** (from step 4)
   - (holding b3): **true** (from previous step)
   - **all preconditions met.**
3. **resulting state:**
   - (handempty)
   - (clear b3)
   - (on b3 b2)
   - (not (clear b2))
   - (not (holding b3))

**step 7: (pick-up b1)**

1. **action and preconditions:**
   - action: pick-up ?ob
   - preconditions: (and (clear ?ob) (ontable ?ob) (handempty))
2. **verification:**
   - (clear b1): **true** (from step 3)
   - (ontable b1): **false** (b1 is not on the table)
   - (handempty): **true** (from previous step)
   - **precondition not met: (ontable b1) is false.**

**conclusion:** **the plan is wrong.** the plan fails at step 7
because b1 is not on the table, violating the precondition 
for the pick-up action. 

Please can you explain the error and fix it.

; Extract plan of the LLM output for the above prompt
;
(unstack b5 b2)
(put-down b5)
(unstack b2 b1)
(put-down b2)
(pick-up b3)
(stack b3 b2)
(pick-up b5)
(stack b5 b3)
(pick-up b2)
(stack b2 b5)
(pick-up b1)
(stack b1 b4)

; LLM output for zero-shot self-critique prompt
;
## Plan Verification:

**Step 1: (unstack b5 b2)**

1. **Action and Preconditions:**
   - Action: unstack (?ob ?underob)
   - Preconditions: (and (on ?ob ?underob) (clear ?ob) (handempty))

2. **Verification:**
   - (on b5 b2): **True** (from initial state)
   - (clear b5): **True** (from initial state)
   - (handempty): **True** (from initial state)
   - **All preconditions met.**

3. **Resulting State:**
   - (on b1 b4)
   - (ontable b4)
   - (on b2 b1)
   - **(holding b5)**
   - **(clear b2)**
   - ontable b3
   - clear b3
   - **(not (on b5 b2))**
   - **(not (handempty))**

**Step 2: (put-down b5)**

1. **Action and Preconditions:**
   - Action: put-down (?ob)
   - Preconditions: (holding ?ob)

2. **Verification:**
   - (holding b5): **True** (from previous step)
   - **Precondition met.**

3. **Resulting State:**
   - (on b1 b4)
   - (ontable b4)
   - (on b2 b1)
   - **(clear b5)**
   - (clear b2)
   - ontable b3
   - clear b3
   - **(ontable b5)**
   - **(handempty)**

**Step 3: (unstack b2 b1)**

1. **Action and Preconditions:**
   - Action: unstack (?ob ?underob)
   - Preconditions: (and (on ?ob ?underob) (clear ?ob) (handempty))

2. **Verification:**
   - (on b2 b1): **True** (from previous step)
   - (clear b2): **True** (from previous step)
   - (handempty): **True** (from previous step)
   - **All preconditions met.**

3. **Resulting State:**
   - (on b1 b4)
   - (ontable b4)
   - **(holding b2)**
   - (clear b5)
   - **(clear b1)**
   - ontable b3
   - clear b3
   - ontable b5
   - **(not (on b2 b1))**
   - **(not (handempty))**

**Step 4: (stack b2 b5)**

1. **Action and Preconditions:**
   - Action: stack (?ob ?underob)
   - Preconditions: (and (clear ?underob) (holding ?ob))

2. **Verification:**
   - (clear b5): **True** (from previous step)
   - (holding b2): **True** (from previous step)
   - **All preconditions met.**

3. **Resulting State:**
   - (on b1 b4)
   - (ontable b4)
   - (clear b1)
   - **(on b2 b5)**
   - **(clear b2)**
   - ontable b3
   - clear b3
   - **(not (clear b5))**
   - ontable b5
   - **(handempty)**

**Step 5: (pick-up b3)**

1. **Action and Preconditions:**
   - Action: pick-up (?ob)
   - Preconditions: (and (clear ?ob) (ontable ?ob) (handempty))

2. **Verification:**
   - (clear b3): **True** (from previous step)
   - (ontable b3): **True** (from previous step)
   - (handempty): **True** (from previous step)
   - **All preconditions met.**

3. **Resulting State:**
   - (on b1 b4)
   - (ontable b4)
   - (clear b1)
   - (on b2 b5)
   - (clear b2)
   - **(holding b3)**
   - **(not (clear b3))**
   - **(not (ontable b3))**
   - (not (clear b5))
   - ontable b5
   - **(not (handempty))**

**Step 6: (stack b3 b2)**

1. **Action and Preconditions:**
   - Action: stack (?ob ?underob)
   - Preconditions: (and (clear ?underob) (holding ?ob))

2. **Verification:**
   - (clear b2): **True** (from previous step)
   - (holding b3): **True** (from previous step)
   - **All preconditions met.**

3. **Resulting State:**
   - (on b1 b4)
   - (ontable b4)
   - (clear b1)
   - **(on b3 b2)**
   - **(clear b3)**
   - **(not (clear b2))**
   - (not (clear b3))
   - (not (ontable b3))
   - (not (clear b5))
   - ontable b5
   - **(handempty)**

**Step 7: (unstack b1 b4)**

1. **Action and Preconditions:**
   - Action: unstack (?ob ?underob)
   - Preconditions: (and (on ?ob ?underob) (clear ?ob) (handempty))

2. **Verification:**
   - (on b1 b4): **True** (from previous step)
   - (clear b1): **True** (from previous step)
   - (handempty): **True** (from previous step)
   - **All preconditions met.**

3. **Resulting State:**
   - **(holding b1)**
   - (ontable b4)
   - **(clear b4)**
   - (on b3 b2)
   - (clear b3)
   - (not (clear b2))
   - (not (clear b3))
   - (not (ontable b3))
   - (not (clear b5))
   - ontable b5
   - **(not (on b1 b4))**
   - **(not (handempty))**

**Step 8: (stack b1 b4)**

1. **Action and Preconditions:**
   - Action: stack (?ob ?underob)
   - Preconditions: (and (clear ?underob) (holding ?ob))

2. **Verification:**
   - (clear b4): **True** (from previous step)
   - (holding b1): **True** (from previous step)
   - **All preconditions met.**

3. **Resulting State:**
   - **(on b1 b4)**
   - **(not (clear b4))**
   - (ontable b4)
   - (on b3 b2)
   - (clear b3)
   - (not (clear b2))
   - (not (clear b3))
   - (not (ontable b3))
   - (not (clear b5))
   - ontable b5
   - **(clear b1)**
   - **(handempty)**

## Conclusion:

The final state includes:
- (on b2 b5)
- (on b3 b2)
- (on b1 b4)

This matches the goal state. Therefore, **the plan is correct**. 

\end{verbatim}
\end{examplebox}

\section{Excerpt from self-improvement trace for Mystery Blocksworld}

\begin{examplebox}[app:sample-trace-my]{Mystery Blocksworld few-shot prompt}
\footnotesize
\begin{verbatim}
The domain definition:
  (define (domain mystery-4ops)
  (:requirements :strips)
(:predicates (province ?x)
             (planet ?x)
             (harmony)
             (pain ?x)
             (craves ?x ?y))

(:action attack
  :parameters (?ob)
  :precondition (and (province ?ob) (planet ?ob) (harmony))
  :effect (and (pain ?ob) (not (province ?ob)) (not (planet ?ob))
               (not (harmony))))

(:action succumb
  :parameters  (?ob)
  :precondition (pain ?ob)
  :effect (and (province ?ob) (harmony) (planet ?ob)
               (not (pain ?ob))))

(:action overcome
  :parameters  (?ob ?underob)
  :precondition (and (province ?underob) (pain ?ob))
  :effect (and (harmony) (province ?ob) (craves ?ob ?underob)
               (not (province ?underob)) (not (pain ?ob))))

(:action feast
  :parameters  (?ob ?underob)
  :precondition (and (craves ?ob ?underob) (province ?ob) (harmony))
  :effect (and (pain ?ob) (province ?underob)
               (not (craves ?ob ?underob)) 
               (not (province ?ob)) (not (harmony)))))

Example of a problem and its solution (plan):
(define (problem MY-rand-7)
(:domain mystery-4ops)
(:objects b6 b3 b1 b7 b4 b2 b5)
(:init
(craves b7 b1)
(planet b6)
(province b3)
(planet b4)
(craves b5 b2)
(province b6)
(province b7)
(planet b3)
(harmony)
(province b5)
(craves b2 b4)
(planet b1)
)
(:goal (and
(craves b3 b2)
(craves b5 b1)
))
)

The plan without formatting:
(feast b7 b1)
(succumb b7)
(feast b5 b2)
(overcome b5 b1)
(attack b3)
(overcome b3 b2)

; rest of few-shot exemplars omitted.
;

And now the problem for you to solve. 
Please solve the following problem:
(define (problem MY-rand-4)
(:domain mystery-4ops)
(:objects b1 b2 b3 b4 )
(:init
(harmony)
(craves b1 b4)
(craves b2 b3)
(planet b3)
(craves b4 b2)
(province b1)
)
(:goal
(and
(craves b1 b3)
(craves b3 b2)
(craves b4 b1))
)
)

Provide only the plan. Do not add numbers, items or any additional text. 
Your plan as plain text without formatting:

; LLM responce
;
(feast b1 b4)
(succumb b1)
(feast b4 b2)
(succumb b4)
(feast b2 b3)
(overcome b2 b1)
(attack b3)
(overcome b3 b2)
(attack b4)
(overcome b4 b1)

; LLM Self-Critique responce (see prompt template 0-shot)
;
## plan verification

**step 1: (feast b1 b4)**

; Verification omitted ...
; 

## conclusion

**the plan is wrong.** the action `(overcome b4 b1)` cannot be executed 
in step 10 because the precondition `(province b1)` is not met. 

Please can you explain the error and fix it.

; LLM responce extract plan only as provided to the self-critique 
;

(feast b1 b4)
(succumb b1)
(feast b4 b2)
(succumb b4)
(feast b2 b3)
(overcome b2 b1)
(attack b3)
(overcome b3 b2)
(attack b4)
(overcome b4 b1)

## Plan Verification

**Initial State:**

; Verification omitted
;

**Goal State:**

* (craves b1 b3)
* (craves b3 b2)
* (craves b4 b1)

**Plan:**

1. **(overcome b1 b4)**

    * **Action Definition:**
        * **Parameters:** (?ob ?underob)
        * **Precondition:** (and (province ?underob) (pain ?ob))
        * **Effect:** (and (harmony) (province ?ob) 
        (craves ?ob ?underob) (not (province ?underob)) 
        (not (pain ?ob)))

    * **Verification:**
        * `?ob = b1`, `?underob = b4`
        * **Precondition Check:**
            * (province b4) - **FALSE** (b4 does not have a province)
            * (pain b1) - **FALSE** (b1 does not have pain)
        * **Preconditions are NOT met.**

    * **Action cannot be applied. The plan is wrong.**

Since the first action of the plan cannot be applied, 
there is no need to evaluate the rest of the plan. 

**Conclusion: The plan is wrong.** 
;
; Rest omitted.
\end{verbatim}
\end{examplebox}

\end{document}